%% file: main.tex
\documentclass[10pt]{article}
\usepackage[letterpaper,margin=1in]{geometry}
\usepackage[T1]{fontenc}
\usepackage{lmodern}
\usepackage{microtype}
\usepackage{amsmath,amssymb}
\usepackage{graphicx}
\usepackage{float}
\usepackage{booktabs}
\usepackage{array}
\usepackage{tabularx}
\usepackage{natbib}
\usepackage{tikz}
\usetikzlibrary{arrows.meta,positioning}
\usepackage{xurl}
\usepackage[colorlinks=true,allcolors=blue,pdfkeywords={3D segmentation, foundation models, 2D-to-3D transfer, open-vocabulary 3D perception, 3D Gaussian Splatting, neural fields, point clouds, semantic scene representations, embodied AI}]{hyperref}
\hypersetup{
  pdftitle={Lift, Associate, and Fuse: A Decision-Centric Framework for 2D-to-3D Foundation Model Transfer},
  pdfauthor={Wentao Sun; Yiping Chen; John S. Zelek; Jonathan Li},
  pdfsubject={Mechanisms, persistent carriers, evaluation protocols, costs, and failure modes in 2D-to-3D foundation model transfer}
}
\title{Lift, Associate, and Fuse: A Decision-Centric Framework for 2D-to-3D Foundation Model Transfer\\\large Auditing Correspondence, Identity, Fusion, and Persistent 3D State}
\author{%
Wentao Sun$^{1}$, Yiping Chen$^{2}$, John S. Zelek$^{1}$, and Jonathan Li$^{1}$\\[0.45em]
\small $^{1}$Department of Systems Design Engineering, University of Waterloo, N2L 3G1 Waterloo, Canada\\
\small $^{2}$School of Geospatial Engineering and Science, Sun Yat-sen University, 519082 Zhuhai, China\\[0.35em]
\small \texttt{wentao.sun@uwaterloo.ca}, \texttt{chenyp79@mail.sysu.edu.cn}\\
\small \texttt{jzelek@uwaterloo.ca}, \texttt{junli@uwaterloo.ca}
}
\date{}
\begin{document}

\newcommand{\PaperKeywords}{%
\par\medskip
\noindent\textbf{Keywords:} 3D segmentation; foundation models; 2D-to-3D transfer; analytical framework; persistent 3D carriers; open-vocabulary 3D perception; 3D Gaussian Splatting; neural fields; embodied AI.
}

\newcommand{\TerminologyCrosswalkTable}{%
\begin{table}[t]
\centering
\caption{Terminology crosswalk used to operationalize decision traces. Terms grouped in one row are common neighbors or aliases, but the boundary column records why they cannot be treated as exact synonyms.}
\label{tab:terminology-crosswalk}
\small
\hbadness=10000
\setlength{\tabcolsep}{3.5pt}
\renewcommand{\arraystretch}{1.12}
\begin{tabularx}{\textwidth}{@{}>{\raggedright\arraybackslash}p{0.19\textwidth}>{\raggedright\arraybackslash}p{0.28\textwidth}>{\raggedright\arraybackslash}X@{}}
\toprule
Concept queried & Common terms encountered & Operational distinction in LAF \\
\midrule
Connecting 2D evidence to 3D & lifting; projection; back-projection; distillation; feature or label transfer & Projection names a geometric operator; lifting is the broader association step; distillation learns a student or carrier; transfer also includes masks, identities, geometry, and labels rather than features alone. \\
Learning and supervision regime & training-free; optimization-free; annotation-free; zero-shot & Training-free may still optimize each scene; optimization-free may use a trained feed-forward model; annotation-free may rely on pseudo-labels; zero-shot concerns held-out concepts or tasks, not the absence of computation or training. \\
Vocabulary and interaction & open-vocabulary; vocabulary-free; open-set; promptable & Open-vocabulary accepts previously unspecified category text; vocabulary-free avoids a fixed enumerated label list but may still constrain language or proposals; promptable only states that user evidence can condition an output. \\
Evaluation domain & rendered-view; image-space; native-3D; primitive-level; proposal-level & Rendered masks score pixels under a renderer; native-3D metrics score points, surfaces, voxels, or Gaussians; proposal-level metrics also depend on the proposal source and matching protocol. \\
Continuous semantic state & semantic field; language field; feature field; embedding field & A semantic field may store class scores; a language field supports text-aligned queries; a feature field can contain self-supervised or task-specific descriptors without language alignment. \\
Retained scene state & persistent carrier; scene representation; semantic memory; map state & Carrier denotes the state actually available to later queries; scene representation can mean geometry alone; semantic memory emphasizes retention or update, and may be an external index or graph rather than the rendered scene. \\
\bottomrule
\end{tabularx}
\end{table}
}

\newcommand{\EvidenceTakeawaysTable}{%
\par\medskip
\refstepcounter{table}%
\label{tab:evidence-takeaways}
\noindent\textbf{Table \thetable.} Framework-derived empirical findings. The middle column identifies the operators and audits that expose each finding; the final column states the condition under which it applies.
\par\smallskip
\small
\hbadness=10000
\setlength{\tabcolsep}{3.5pt}
\renewcommand{\arraystretch}{1.12}
\noindent
\begin{tabularx}{\textwidth}{@{}>{\raggedright\arraybackslash}p{0.32\textwidth}>{\raggedright\arraybackslash}p{0.20\textwidth}>{\raggedright\arraybackslash}X@{}}
\toprule
Empirical finding & LAF evidence & Scope condition \\
\midrule
Association is not identity reconciliation. & Associate and Reconcile traces; Tables 1 and 6 & Applies to hard projection, renderer-mediated support, and learned correspondence: correct support still does not determine whether observations share an entity, label, or granularity. \\
Training-free does not imply low end-to-end cost. & Stage-scoped cost audit; Table 5 & Reconstruction, 2D foundation-model inference, association, reconciliation, storage, and query costs must be included; the finding concerns the complete deployment ledger, not one downstream module. \\
Rendered-view scores do not establish native-3D topology. & Output-contract audit; Table 4 & Cross-method comparison is valid only under the same rendering, extraction, geometry, prompt, proposal, and oracle protocol; image agreement can hide disconnected or incomplete 3D support. \\
Persistent carriers define both query capability and correction boundaries. & Carrier contract; Table 3 & The consequence depends on what is retained: hard IDs, continuous features, object inventories, relations, provenance, and uncertainty permit different retrieval and revision operations. \\
Open-vocabulary capability is stage- and interface-specific. & Vocabulary-timing trace; Table 2 & Text access at the 2D teacher, carrier, classifier, or query interface is not equivalent; a new query may require rerunning upstream models even when the reported output is open-vocabulary. \\
\bottomrule
\end{tabularx}
\par\medskip
}

\newcommand{\PipelineAuditTable}{%
\begin{table}[t]
\centering
\caption{Mechanism-first audit of the five transfer stages. The stages describe decisions, not mandatory software modules; one model may implement several jointly.}
\label{tab:pipeline-audit}
\small
\hbadness=10000
\setlength{\tabcolsep}{4pt}
\renewcommand{\arraystretch}{1.13}
\begin{tabularx}{\textwidth}{@{}>{\raggedright\arraybackslash}p{0.12\textwidth}>{\raggedright\arraybackslash}p{0.22\textwidth}>{\raggedright\arraybackslash}p{0.26\textwidth}>{\raggedright\arraybackslash}X@{}}
\toprule
Stage & Decision being made & Representative mechanisms & First irreversible risk \\
\midrule
Generate & What evidence is exposed, at which scale, and with what confidence? & SAM masks and tracks, CLIP/OpenSeg features, DINO structure, generated captions \citep{peng_2023_openscene,ye_2024_gaussian_grouping,ding_2023_pla} & A missing object, part, or label cannot be reconstructed after the source evidence is discarded. \\
Associate & Which point, voxel, ray sample, or Gaussian may receive each observation? & Calibrated projection, NeRF transmittance, Gaussian contribution weights, learned point maps \citep{kerr_2023_lerf,shi_2024_legaussians,gong_2026_ov3r} & Pose, depth, visibility, or renderer errors move otherwise correct evidence to the wrong surface. \\
Reconcile & Which observations share identity, meaning, or granularity? & Tracking, overlap graphs, matching, clustering, hierarchy and scale conditioning \citep{yin_2024_sai3d,yan_2024_maskclustering,he_2024_ultrametric} & A false merge, split, or parent relation becomes the topology inherited by later queries. \\
Fuse & How are repeated and conflicting observations combined? & Weighted means, robust selection, learned fields, sparse codes, external indices \citep{jatavallabhula_2023_conceptfusion,shi_2024_legaussians,budimir_2026_scoup} & Averaging or compression can erase minority evidence while making the result appear confident. \\
Persist/Query & What state remains, and what can a later query return or revise? & Point features, neural fields, Gaussian attributes, object sets, scene graphs, temporal memories \citep{peng_2023_openscene,wu_2024_opengaussian,xie_2026_relags} & The stored carrier fixes the available query surface, update rule, and recovery boundary. \\
\bottomrule
\end{tabularx}
\end{table}
}

\newcommand{\CarrierComparisonTable}{%
\begin{table}[t]
\centering
\caption{Persistent carrier families and their query and correction contracts. ``Natural output'' denotes the representation returned without an additional conversion protocol.}
\label{tab:carrier-comparison}
\small
\hbadness=10000
\setlength{\tabcolsep}{4pt}
\renewcommand{\arraystretch}{1.13}
\begin{tabularx}{\textwidth}{@{}>{\raggedright\arraybackslash}p{0.16\textwidth}>{\raggedright\arraybackslash}p{0.20\textwidth}>{\raggedright\arraybackslash}p{0.27\textwidth}>{\raggedright\arraybackslash}X@{}}
\toprule
Carrier & Natural output & Main advantage & Correction boundary \\
\midrule
Points / voxels / meshes & Native 3D labels, masks, or features & Direct geometric extraction and compatibility with standard 3D benchmarks \citep{peng_2023_openscene,nguyen_2024_open3dis} & Quantization, visibility filtering, and hard proposal topology can be difficult to undo. \\
Neural fields & Rendered relevance, affinity, or mask values & Continuous spatial queries and differentiable visibility \citep{kerr_2023_lerf,kim_2024_garfield} & Native 3D extraction requires sampling and thresholds; corrections may require scene optimization. \\
Gaussian attributes & Rendered and primitive-level responses & Explicit editable primitives with fast differentiable rendering \citep{wu_2024_opengaussian,shi_2024_legaussians} & Semantics inherit reconstruction geometry, alpha compositing, and per-Gaussian storage choices. \\
Object sets & Ranked objects, masks, and descriptors & Compact repeated retrieval and object-level updates \citep{takmaz_2023_openmask3d,lu_2023_ovir3d} & Objects absent or wrongly merged during construction are unavailable to later language queries. \\
Graphs / hierarchies & Nodes, relations, paths, and multiscale supports & Relational and parent-aware reasoning beyond unary segmentation \citep{werby_2024_hovsg,wang_2025_octree_graph} & One node or parent error can alter retrieval, relations, and planning simultaneously. \\
Temporal memories & Identity- and time-conditioned masks or states & Online updates, reappearance handling, and spatio-temporal queries \citep{xu_2025_sam4d,meng_2026_st4rsplat} & Identity switches and stale evidence become persistent state-transition errors. \\
\bottomrule
\end{tabularx}
\end{table}
}

\newcommand{\EvaluationProtocolTable}{%
\begin{table}[t]
\centering
\caption{Evaluation families that should remain separate unless the same conversion and oracle protocol is applied to every method.}
\label{tab:evaluation-families}
\small
\hbadness=10000
\setlength{\tabcolsep}{4pt}
\renewcommand{\arraystretch}{1.13}
\begin{tabularx}{\textwidth}{@{}>{\raggedright\arraybackslash}p{0.17\textwidth}>{\raggedright\arraybackslash}p{0.20\textwidth}>{\raggedright\arraybackslash}p{0.25\textwidth}>{\raggedright\arraybackslash}X@{}}
\toprule
Evaluation family & Prediction unit & Compatible evidence & Required disclosure \\
\midrule
Rendered localization & Pixels in held-out or prompted views & mIoU, boundary IoU, relevancy localization under a shared renderer and view set & Rendering backbone, test views, scale selection, and whether native geometry is evaluated. \\
Native semantic segmentation & Points, voxels, mesh faces, or Gaussians & mIoU or class accuracy on the same geometry, split, and label mapping & Geometry source, ignored labels, vocabulary access, and feature-to-label conversion. \\
Native instance / panoptic segmentation & 3D proposals or instance masks & AP, AP50/AP25, PQ, recognition and segmentation quality under the same proposal protocol & Proposal source, class-agnostic versus semantic scoring, oracle masks, and overlap threshold. \\
Part and hierarchical segmentation & Parts, trees, or threshold-selected partitions & Part IoU, hierarchical consistency, and stability under shared object and prompt sets & Part taxonomy, orientation, granularity control, prompt visibility, and parent-child ground truth. \\
Online / dynamic segmentation & Frame identities, tracks, or space-time supports & Frame AP, temporal IoU, identity consistency, update and recovery measures & Sensor stream, observation order, target identity access, disappearance, and correction protocol. \\
Relational / downstream tasks & Graph edges, retrievals, paths, edits, or actions & Relation recall, retrieval accuracy, navigation success, or edit quality within the same task contract & Upstream oracle nodes, language model calls, map state, task policy, and whether segmentation is scored separately. \\
\bottomrule
\end{tabularx}
\end{table}
}

\newcommand{\CostLedgerTable}{%
\begin{table}[t]
\centering
\caption{Full cost ledger for separating one-time construction, persistent storage, and marginal query cost. NR should be reported when a component is not measured.}
\label{tab:cost-ledger}
\small
\hbadness=10000
\setlength{\tabcolsep}{4pt}
\renewcommand{\arraystretch}{1.13}
\begin{tabularx}{\textwidth}{@{}>{\raggedright\arraybackslash}p{0.16\textwidth}>{\raggedright\arraybackslash}p{0.28\textwidth}>{\raggedright\arraybackslash}p{0.18\textwidth}>{\raggedright\arraybackslash}X@{}}
\toprule
Ledger stage & What must be counted & Cost scope & Common omission \\
\midrule
Geometry & RGB-D fusion, SfM, NeRF/3DGS optimization, or feed-forward reconstruction & Per scene or per stream & Calling a method reconstruction-free when calibration or pretrained geometry is still required. \\
2D evidence & Mask generation, tracking, detection, captioning, and feature extraction over all views & Per image, video, or prompt & Reporting only the downstream 3D module while omitting foundation-model calls. \\
Association & Projection, visibility tests, inverse rendering, view selection, and temporary point-map alignment & Per observation or scene & Treating correspondence as free preprocessing. \\
Reconciliation & Matching, graph construction, clustering, hierarchy recovery, and identity maintenance & Per scene or update & Excluding iterative merge, tracking, or graph costs from ``fusion'' time. \\
Carrier construction & Distillation, field optimization, code learning, or cross-scene training & Per scene or amortized training & Comparing a feed-forward pass with per-scene fitting while omitting pretraining accelerator-hours. \\
Persistence & Carrier parameters, external indices, raw evidence, and transient peak memory & Bytes per scene and peak device memory & Reporting model size without scene state, or scene state without retained images and indices. \\
Query / update & First query, repeated query, language-agent calls, rendering, extraction, correction, and reindexing & Per query or state change & Quoting renderer FPS when reasoning, refinement, or detector calls dominate end-to-end latency. \\
\bottomrule
\end{tabularx}
\end{table}
}

\newcommand{\FailureDiagnosticTable}{%
\begin{table}[t]
\centering
\caption{Failure sources, their propagation paths, and diagnostics that identify the first irreversible information loss.}
\label{tab:failure-diagnostics}
\small
\hbadness=10000
\setlength{\tabcolsep}{4pt}
\renewcommand{\arraystretch}{1.13}
\begin{tabularx}{\textwidth}{@{}>{\raggedright\arraybackslash}p{0.16\textwidth}>{\raggedright\arraybackslash}p{0.27\textwidth}>{\raggedright\arraybackslash}p{0.25\textwidth}>{\raggedright\arraybackslash}X@{}}
\toprule
Failure source & Propagation into the carrier & Diagnostic & Recovery requirement \\
\midrule
Geometry / visibility & Correct evidence is assigned to an occluder, floater, or wrong surface & Pose/depth perturbation, oracle geometry, and visible-only support precision/recall & Reprojection audit and local reassociation without rebuilding unrelated state. \\
Proposal topology & Missing, merged, or over-fragmented masks define an incomplete object inventory & Proposal-oracle versus classifier-oracle evaluation; small-object and part recall & New proposal admission and reversible split/merge operations. \\
Identity & Tracker switches or false cross-view matches become persistent object state & Observation-order replay, injected identity switches, disappearance and reappearance tests & Merge provenance, confidence, rollback, and contradictory-evidence handling. \\
Semantic conflict & Correlated teacher errors or averaging produce confident wrong labels & Per-teacher ablation, calibration, counterfactual context, and disagreement retention & Source-aware fusion, abstention, and relabeling without changing geometry. \\
Granularity / hierarchy & One scale or bottleneck edge changes many part and parent decisions & Scale and threshold sweeps, parent perturbation, and partition stability & Query-dependent granularity and reversible hierarchy edits. \\
Compression & Quantization, Top-K atoms, or hard IDs erase rare but useful evidence & Query-after-compression, rare-query recall, code occupancy, and storage-accuracy curves & Expandable codes, retained provenance, or external evidence lookup. \\
Dynamic updates & Stale features, identity drift, or unsynchronized indices survive scene changes & Controlled lifecycle changes, delayed contradictions, and transactional consistency tests & Local invalidation, state versioning, and synchronized map/index updates. \\
Relations / agents & Missing nodes and plausible language errors alter retrieval, planning, or edits & Node/edge oracles, language-call ablations, evidence traces, and corrective-query tests & Exposed intermediate decisions and correction without full scene reconstruction. \\
\bottomrule
\end{tabularx}
\end{table}
}
\maketitle
\begin{abstract}
Methods that transfer predictions from two-dimensional foundation models into three-dimensional segmentation are commonly grouped by task or representation. Those groupings obscure the decisions that determine whether a system remains coherent across views: where image evidence is grounded, when observations become one identity, how semantic and granularity conflicts are handled, which information is fused, and what state survives for later queries. We introduce \textbf{Lift, Associate, and Fuse (LAF)}, a decision-centric framework that represents a transfer system as five operators: \textbf{Generate, Associate, Reconcile, Fuse, and Persist/Query}. LAF defines an explicit contract for the persistent carrier---its spatial support, semantic state, identity state, uncertainty, provenance, and supported operations---and identifies the first stage at which discarded evidence becomes unrecoverable. We operationalize the framework as a structured audit protocol and apply it to 161 systems available through 7 August 2026, spanning point-, field-, Gaussian-, object-, graph-, and memory-based carriers. Representation, temporal, relational, and feed-forward stress tests required no additional analytical stage after the final confirmation pass. The resulting decision traces expose four recurring properties: association does not establish identity; carrier design fixes both the query interface and correction boundary; rendered-view, native-3D, and proposal-level evaluations are not interchangeable; and qualifiers such as \emph{training-free}, \emph{real-time}, \emph{open-vocabulary}, and \emph{generalizable} are meaningful only when attached to a stage and a complete cost ledger. LAF therefore supplies a representation-neutral method for comparing existing systems, diagnosing irreversible failures, and specifying revisable 3D perception for future agents.

\noindent\textbf{Webpage:} \href{https://w27sun.github.io/LAFSum/}{\texttt{https://w27sun.github.io/LAFSum/}}

\PaperKeywords

\end{abstract}
\input{overview_figure}

\section{Introduction}

Two-dimensional foundation models have changed both what segmentation systems can recognize and how users can prompt them. Mask generators propose objects and parts without task-specific retraining; vision-language encoders connect regions to category names and descriptions; self-supervised encoders provide dense structure; trackers propagate identities; and monocular models estimate geometry. Yet correct predictions in individual images do not automatically form a coherent 3D segmentation. A 3D system must still determine which surface produced each pixel, whether observations from different views describe the same entity, how part and object scales coexist, how contradictory evidence is combined, and what information remains after the images are discarded.

Existing classifications usually begin with a task label---semantic, instance, panoptic, part, open-vocabulary, or promptable segmentation---or a representation label---point cloud, neural field, or 3D Gaussian Splatting (3DGS). Those labels are useful for retrieval but weak for diagnosis. Two systems using the same mask generator and the same 3DGS reconstruction may establish identity at different times, retain different evidence, expose different query interfaces, and fail under different perturbations. Conversely, a point system and a neural field may implement the same decision sequence through different numerical operators.

The phrase \emph{2D-to-3D lifting} makes the problem appear to be one projection. In practice it may denote calibrated back-projection through depth, supervision through NeRF transmittance, accumulation of Gaussian rendering contributions, learned cross-view attention, or registration through a temporary reconstruction. More importantly, it conflates three decisions. \textbf{Association} locates evidence on 3D support. \textbf{Reconciliation} determines whether observations share an identity, semantic label, or part--whole relation. \textbf{Fusion} constructs reusable state. Correct association does not establish identity, and feature averaging does not specify what a later user or agent can query or correct.

We introduce \textbf{Lift, Associate, and Fuse (LAF)}, a decision-centric framework with five analytical operators:

1. \textbf{Generate:} expose masks, embeddings, identities, language, confidence, or geometry from 2D models. 2. \textbf{Associate:} connect that evidence to explicit or implicit 3D support. 3. \textbf{Reconcile:} resolve or retain cross-view identity, semantic disagreement, and granularity. 4. \textbf{Fuse:} convert repeated and conflicting observations into reusable state. 5. \textbf{Persist/Query:} render, retrieve, edit, update, or reason over the resulting carrier.

The operators describe decisions rather than mandatory software modules. One objective may implement several operators jointly, and one operator may be revisited by an online feedback loop. This abstraction makes a system auditable because it records what information each operator receives, what it discards, and which downstream operations remain possible. Figure 1 shows the decision flow and its update loop. Table 1 states the corresponding audit questions and the first representative loss that can become irreversible at each stage.

\PipelineAuditTable

LAF is built around a \textbf{persistent carrier contract}. Point-based systems may retain per-point labels or language features \citep{peng_2023_openscene,nguyen_2024_open3dis}; neural fields may retain continuous relevance, affinity, or mask functions \citep{kerr_2023_lerf,engelmann_2024_opennerf}; Gaussian systems may retain per-primitive semantics, hard memberships, sparse codes, decoder correlations, or relation graphs \citep{ye_2024_gaussian_grouping,shen_2024_flashsplat,zhang_2025_infogs}. These outputs are not interchangeable even when they render similar colored masks. Their support, identity state, uncertainty, provenance, and permitted operations determine what a future query can recover and what a correction can revise.

The framework makes four contributions.

\textbf{A compositional decision model.} We formalize Generate, Associate, Reconcile, Fuse, and Persist/Query as operators whose intermediate states form a decision trace. The decomposition separates support correspondence from entity identity and exposes when vocabulary or granularity enters the pipeline.

\textbf{A carrier and irreversibility contract.} We characterize persistent state by spatial support, semantic state, identity, uncertainty, provenance, and supported operations. This contract identifies the first stage after which a discarded hypothesis cannot be recovered without rerunning an upstream model or rebuilding the scene.

\textbf{A representation-neutral audit protocol.} We operationalize LAF on 161 systems available through 7 August 2026. The corpus spans six carrier families and deliberately stresses calibrated, renderer-mediated, learned, temporary, online, relational, dynamic, and feed-forward mechanisms. The final confirmation pass required no additional analytical stage.

\textbf{Framework-derived empirical findings.} The decision traces show that association and identity are distinct; carrier design fixes the query and correction boundary; rendered-view, native-3D, and proposal-level scores answer different questions; and efficiency or vocabulary qualifiers must be scoped to a stage and a complete cost ledger. These findings lead directly to reporting rules and design requirements for revisable agentic 3D perception.

Section 2 formalizes LAF and describes its construction and validation protocol. Section 3 introduces the geometric and rendering operators used by concrete implementations. Sections 4--8 instantiate the five stages. Sections 9--10 apply the framework to evaluation, cost, and failure propagation. Sections 11--12 extend the carrier contract to agentic perception and identify framework-derived research problems; Section 13 concludes.

\section{The LAF Framework and Validation Protocol}

\subsection{Operator model and decision trace}

Let $I_v$ denote an image observed from view $v$, $\pi_v$ its camera model, $\mathcal G$ the available explicit or implicit geometry, and $q$ a downstream query. LAF writes a system as the composition

\[
e_v = G(I_v;q), \qquad
s_v = A(e_v,\mathcal G,\pi_v), \qquad
h = R(\{s_v\}_{v=1}^{V}), \qquad
C = F(h), \qquad
y = Q(C,q).
\]

$G$ exposes image evidence; $A$ assigns support in 3D; $R$ constructs or preserves identity, semantic, and granularity hypotheses; $F$ materializes reusable state; and $Q$ reads or changes that state. A method's \emph{decision trace} is

\[
\tau(M)=\langle G,A,R,F,C,Q\rangle,
\]

augmented with the supervision available to each operator, the state discarded after it, and the costs charged to it. The trace is valid even when operators share parameters or are optimized jointly: the question is which decision is implemented, not whether the code contains a correspondingly named module.

This factorization prevents three frequent category errors. First, a nonzero support weight $s_v$ says where evidence may belong but not whether two observations share an entity. Second, a common embedding space does not imply a persistent object identity. Third, a rendered response is an observation of $C$ through $Q$, not necessarily a native segmentation of $\mathcal G$.

\subsection{Persistent carrier contract}

We define the carrier by the contract

\[
C = (\mathcal S,\mathcal Z,\mathcal I,\mathcal U,\mathcal P,\mathcal O),
\]

where $\mathcal S$ is spatial support, $\mathcal Z$ is semantic or feature state, $\mathcal I$ is identity and topology, $\mathcal U$ is retained uncertainty, $\mathcal P$ is provenance, and $\mathcal O$ is the set of supported operations. $\mathcal O$ may include native extraction, novel-view rendering, text retrieval, object selection, split/merge correction, online update, relational reasoning, or planning. Two methods with the same representation are different carriers when these contracts differ; two different representations can be operationally comparable when they expose the same contract under an explicit conversion.

The contract also separates \emph{state} from \emph{service}. A query-time VLM may provide rich language answers while $C$ retains only geometry and hard IDs; conversely, a dense language field may answer text queries without storing explicit object identity. The cost and correction consequences follow the retained tuple, not the name of the external model.

\subsection{First irreversible loss}

Let $\mathcal H_j$ be the hypotheses recoverable from the state retained after stage $j$. A loss at stage $j$ is irreversible for a downstream query family when a relevant hypothesis $h$ is absent from $\mathcal H_j$ and no later operator receives the discarded evidence needed to reconstruct it. The \emph{first irreversible loss} is the earliest such stage. This definition is query-relative: discarding per-view mask provenance may be harmless for one fixed semantic benchmark but irreversible for a later request to split a false merge or explain a label.

The lower path of Figure 1 instantiates this definition. A missing proposal limits $\mathcal H_G$; a visibility error moves support before reconciliation; a false merge changes $\mathcal I$; lossy fusion removes alternatives from $\mathcal U$ or $\mathcal P$; and a narrow persistence interface limits $\mathcal O$. Diagnosing the first irreversible loss prevents a downstream language model or smoother from being credited with information that the carrier no longer contains.

\begin{figure}[H]
\centering
\includegraphics[width=\textwidth]{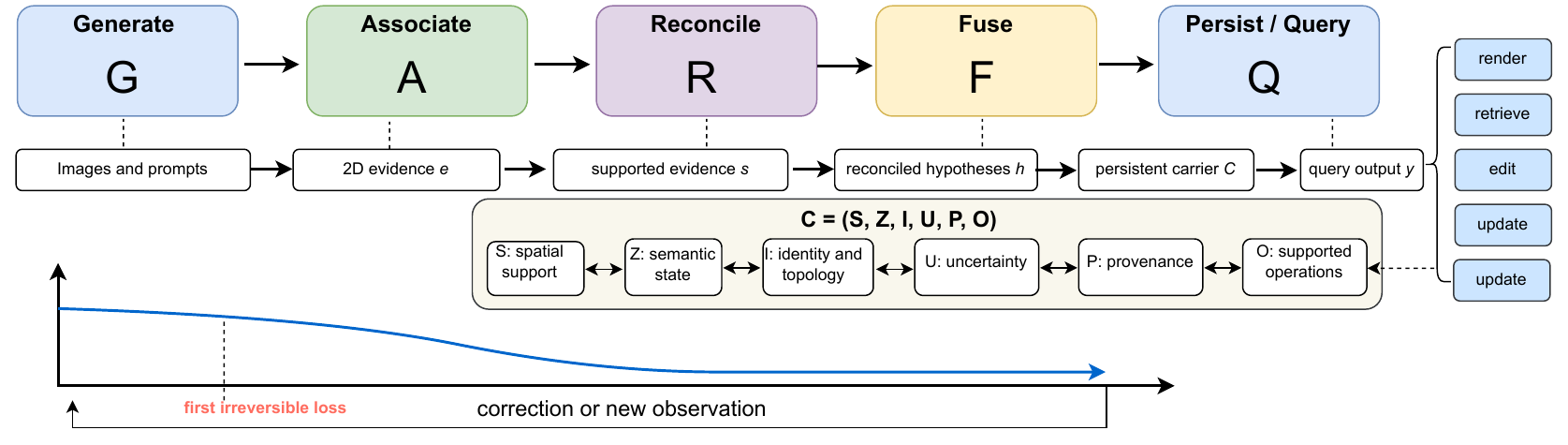}
\caption{Formal LAF decision trace and persistent-carrier contract. Each operator transforms or constrains the hypotheses available to the next stage. The carrier is specified by spatial support $\mathcal S$, semantic state $\mathcal Z$, identity and topology $\mathcal I$, uncertainty $\mathcal U$, provenance $\mathcal P$, and supported operations $\mathcal O$. The lower band depicts progressive loss of recoverable alternatives, while the feedback path represents correction or additional observation.}
\label{fig:laf-contract}
\end{figure}

Figure~\ref{fig:laf-contract} combines the operator trace, carrier tuple, query interface, and recovery path in one formal view. The narrowing band denotes the set of hypotheses that remain recoverable as decisions accumulate; its width is conceptual rather than a measured information quantity. The warning marker illustrates that the first irreversible loss is query-dependent and may occur at a different operator for a different system or downstream request.

\subsection{Framework construction corpus}

We constructed and stress-tested LAF on methods that transfer knowledge from a broadly pretrained 2D vision or vision-language model into a material 3D segmentation or persistent 3D label state. Eligible outputs include semantic, instance, panoptic, part, open-vocabulary, and promptable segmentation on points, voxels, meshes, neural fields, 3D Gaussians, object structures, and temporal memories. Transferred evidence may be a mask, label, language-aligned embedding, self-supervised feature, track identity, detector output, monocular geometry estimate, or generated description.

A method enters the validation corpus when it satisfies four conditions: (1) it produces, supervises, evaluates, or directly enables a material 3D segmentation or persistent 3D label; (2) it transfers knowledge from a broadly pretrained 2D model; (3) it contains a substantive association, reconciliation, fusion, or persistence decision; and (4) its central claims can be checked against primary-source evidence. Purely 2D segmentation, conventional supervised 3D segmentation without relevant transfer, and reconstruction or editing work with only incidental segmentation are outside the contract. PointGauss is retained as a boundary case because it operates on Gaussian geometry but its segmentation mechanism is a native-3D trained decoder rather than transferred image evidence \citep{sun_2025_pointgauss}.

The unit of analysis is a method rather than a PDF version. Discovery combined transfer terms (\texttt{lift}, \texttt{project}, \texttt{distill}, \texttt{2D-to-3D}, \texttt{multi-view}), representation terms (\texttt{point cloud}, \texttt{NeRF}, \texttt{neural field}, \texttt{Gaussian Splatting}), segmentation tasks, and source-model families including SAM, CLIP, DINO, detectors, grounders, trackers, and monocular geometry models. Backward references, forward citations, and neighboring mechanisms expanded the seed set.

The cutoff is 2026-08-07. The final snapshot contains 166 indexed methods: 161 satisfy the framework contract, four are retained as boundary or adjacent context, and one is excluded after full-text screening. The evidence set contains 165 parsed PDFs comprising 2,277 pages. OV3DSeg-VGGT is the sole mechanism-only publisher-HTML exception, and no numerical result from that method enters the analysis.

\subsection{Audit procedure}

Each included method is converted into a trace $\tau(M)$ through five passes. The first records the generated evidence and its vocabulary, scale, confidence, and provenance. The second reconstructs the support operator and its visibility assumptions. The third identifies when identity, semantic conflict, and granularity become fixed or remain revisable. The fourth records the fusion rule and which alternatives are discarded. The fifth instantiates the carrier tuple and lists native queries, conversions, updates, and corrections. A separate ledger records geometry construction, 2D inference, association, reconciliation, carrier construction, storage, and first and repeated queries.

Claims are tied to a page, section, equation, figure, table, or appendix in the primary source. Abstracts support discovery but not experimental claims. Method diagrams and result tables are inspected visually because text extraction can lose symbols, alignment, and multi-column order. Central claims receive a second fact check before they are treated as stable. We do not pool accuracy across incompatible geometry, vocabulary, proposal, prompt, output, or denominator protocols.

\subsection{Framework stress tests and observed coverage}

We evaluated analytical coverage rather than predictive accuracy. Four stress tests were applied.

\textbf{Representation stress.} The same trace schema was applied to points/voxels/meshes, neural fields, Gaussian attributes, object sets, graphs/hierarchies, and temporal memories. No carrier required a sixth decision stage; differences appeared in the carrier tuple and in how operators were implemented.

\textbf{Mechanism stress.} Calibrated projection, renderer-mediated weights, learned correspondence, and temporary-carrier registration all instantiate $A$ without forcing identity into the association operator. Tracker-first, clustering, matching, prototype, and joint-optimization systems instantiate distinct $R$ operators.

\textbf{State-evolution stress.} Per-scene fields, feed-forward reconstruction, online maps, dynamic Gaussians, relational graphs, and query-time agents can all be represented by allowing feedback from $Q$ to $G$, $A$, or $R$ and by versioning $C$. The feedback edge changes when a decision is revisited, not the five decisions that must be made.

\textbf{Boundary and saturation stress.} Native-3D promptable models, reconstruction-only systems, rendered-only semantics, and relation graphs without material segmentation were used as negative or adjacent cases. Representation-focused searches were followed by non-Gaussian, venue-neutral, online, relational, dynamic, and feed-forward confirmation rounds. The final confirmation pass produced neither a missing canonical mechanism nor a new core operator. This is evidence of analytical saturation at the cutoff, not proof that future work cannot extend the framework.

Four cross-representation traces illustrate what the stress test distinguishes. OpenScene generates dense language features, associates them by calibrated depth-aware projection, performs no instance reconciliation, fuses them by point averaging and distillation, and persists point features \citep{peng_2023_openscene}. SA3D uses renderer weights for association and closes a query-to-generation feedback loop through self-prompting while persisting a soft mask field \citep{cen_2023_sa3d}. Gaussian Grouping fixes track identity before learning an identity-bearing Gaussian field \citep{ye_2024_gaussian_grouping}. OGScene3D revisits reconciliation and persistence as observations arrive, storing confidence-bearing Gaussian labels and a progressively revised graph \citep{zhu_2026_ogscene3d}. A representation-only taxonomy groups the last three as field or Gaussian methods; LAF exposes their different identity times, update paths, and correction boundaries.

\subsection{Validity and limitations}

The framework is an analytical instrument, not a claim that software always contains five separable modules. Its validity depends on faithful reconstruction of decisions from published evidence. Rapid publication and inconsistent naming can still cause omissions, and incomplete runtime reporting leaves parts of the cost contract qualitative. The current validation establishes representation and mechanism coverage through structured application and saturation checks; it does not report formal inter-annotator reliability, because the historical evidence cards were not all produced under a prospective double-coding design. A future registered coding study should measure agreement on stage boundaries, identity establishment time, and carrier operations.

The framework deliberately focuses on systems with a material 3D segmentation pathway. Pure 2D cross-view consistency, 3D referring expressions without 2D transfer, generic reconstruction, and scene graphs without segmentation enter only as boundary tests. Internal evidence cards and screening controls support fact checking; the public research claims are contained in this article and its presentation website rather than a separate public knowledge base.

\section{Operational Definitions and Correspondence Operators}

The reviewed systems move several different objects across representations: colors, masks, language embeddings, self-supervised features, track identities, affinities, and uncertainty. Calling all of them ``features'' obscures which operations are valid. We therefore introduce notation around evidence, support, and carrier state before comparing implementations.

\subsection{Images, geometry, and evidence}

Let image $I_v$ be observed from view $v$, with camera intrinsics $K_v$ and pose $T_v$. A 2D model produces evidence $E_v(u)$ at pixel $u$. Depending on the method, $E_v$ may be a continuous vector, a class-score distribution, a binary mask membership, a discrete track ID, or a set of nested regions. The 3D representation contains elements $g_i$: points, voxels, surface samples, neural-field locations, Gaussian primitives, or object nodes.

Association produces a support weight $a_{viu}$ describing how strongly pixel $u$ in view $v$ supports element $g_i$. Reconciliation then reasons over the collection of supported observations, for example by matching mask IDs, clustering affinities, selecting one scale, or preserving several overlapping hypotheses. Fusion creates carrier state $z_i$, a hard membership $m_{ik}$, a field parameterization, or a graph. A query $q$ finally maps that persistent state to an output $Y(q)$. This notation separates a physical correspondence $a$ from semantic state $z$: a correct projection does not imply that two observations share an object identity or label.

\subsection{Calibrated projection and visibility}

For explicit geometry, a point $x_i$ is transformed and projected as $\tilde{u}_{vi} = \pi(K_v T_v x_i)$. A valid correspondence additionally requires positive depth, image bounds, and a visibility test comparing projected depth with the measured or rendered depth at that pixel. Hard lifting sets $a_{viu}$ to one for the accepted pixel and zero otherwise. Soft variants weight evidence by depth residual, view angle, object visibility, or confidence.

Projection appears simple because its geometry is explicit, but its negative evidence policy is consequential. If an occluded point is treated as visible, a background pixel becomes false supervision; if a tolerance is too strict, valid evidence disappears. OpenScene, ConceptFusion, SAM3D, and Open3DIS all use calibrated association, but they lift different objects and fuse them under different visibility assumptions \citep{peng_2023_openscene,jatavallabhula_2023_conceptfusion,yang_2023_sam3d,nguyen_2024_open3dis}.

\subsection{NeRF volume rendering as soft association}

A neural radiance field predicts density $\sigma(t)$ and attributes along a camera ray $r(t)$. With ordered samples, the contribution of sample $i$ is

\[
w_i = T_i \alpha_i, \qquad
T_i = \prod_{j<i}(1-\alpha_j), \qquad
\alpha_i = 1-\exp(-\sigma_i\delta_i).
\]

Color and semantic attributes are rendered by weighted accumulation, for example $\hat z(r)=\sum_i w_i z_i$. The same weights can send 2D supervision backward into a feature or mask field. LERF stores scale-conditioned language relevance, SA3D optimizes a prompted mask field, and OpenNeRF distills localized OpenSeg features \citep{kerr_2023_lerf,cen_2023_sa3d,engelmann_2024_opennerf}. The renderer supplies differentiable, partial-visibility association, but the weights are only as correct as the learned density.

Rendered feature matching is not the only supervision path. Rethinking OVRF applies semantic loss directly to sampled 3D points and then transfers the learned NeRF language field into a 3DGS carrier \citep{lee_2025_rethinking_ovrf}. This separates the representation used to establish semantics from the representation used for fast rendering, but introduces a second correspondence problem across carriers.

\subsection{Gaussian alpha compositing}

In 3DGS, primitive $g_i$ has a 3D mean, covariance, opacity, appearance, and optionally semantic attributes. Projection yields an elliptical footprint with per-pixel opacity $\alpha_i(u)$. After depth sorting, its image contribution has the analogous form

\[
c_i(u) = \alpha_i(u)\prod_{j<i}(1-\alpha_j(u)).
\]

Methods use $c_i(u)$ in several non-equivalent ways. Gaussian Grouping renders learned identity codes; LangSplat renders compressed language features; FlashSplat accumulates contributions into a closed-form membership vote; Dr. Splat registers features only to dominant primitives; and InfoGS assigns coarse labels by the maximum contribution before shaping decoder correlations \citep{ye_2024_gaussian_grouping,qin_2024_langsplat,shen_2024_flashsplat,kim_2025_dr_splat,zhang_2025_infogs}. Ray intersection, projected footprint overlap, maximum contribution, and marginal contribution are therefore different association policies even when all are described as Gaussian lifting.

Renderer support can also be coupled to an external feature parameterization. FMGS renders a multi-resolution hash-encoded CLIP/DINO field through Gaussian support rather than storing the full feature only as a free vector on each splat \citep{zuo_2025_fmgs}. Consequently, geometric carrier, rendering support and semantic storage should be recorded as separate axes.

\subsection{Identity, semantics, and granularity}

An identity variable answers whether observations belong to the same physical entity; a semantic variable answers what the entity means; granularity determines whether the entity is a part, object, or group. These variables may be coupled but should not be collapsed. Tracker IDs can reconcile an object before 3D learning, a learned code can establish identity during field optimization, and geometry-based merging can decide it afterward. A language field can remain useful without discrete instances, while a hard instance carrier can remain class-agnostic until query time.

Granularity may be fixed by the 2D masks, selected among discrete levels, conditioned on physical scale, or stored as a hierarchy or relation graph. The choice determines whether later queries can revise a partition. A hard label discards alternative memberships; an affinity graph preserves more options but costs memory and requires clustering. Throughout this work, \emph{carrier} denotes the persistent state available after construction, while \emph{evidence} denotes the observations used to create it.

\subsection{Training and vocabulary regimes}

Per-scene optimization fits a carrier to one reconstructed scene. Cross-scene training instead learns parameters reused on new scenes, while online systems revise state as frames arrive. These regimes move cost between construction and deployment and should not be ranked by query latency alone. Vocabulary timing is independent: labels may be fixed before training, encoded as language-aligned features during construction, or attached to class-agnostic identities after geometry is frozen. We reserve \emph{open-vocabulary} for interfaces that accept previously unspecified category text, and qualify whether the stored representation can actually respond without rerunning upstream 2D models.

\subsection{Terminology crosswalk}

Terminology varies across point-cloud, neural-field, Gaussian, and embodied-perception communities. Table 2 records common aliases while preserving distinctions that affect mechanism, supervision, evaluation, or deployment cost. The entries are operational definitions for constructing a trace, not declarations that all terms in one row are interchangeable. Throughout the framework, the more specific term is used whenever the evidence permits it.

\TerminologyCrosswalkTable

\section{Generate: What Comes from 2D?}

The generation stage determines the information ceiling of the entire pipeline. A 3D carrier can reconcile inconsistent observations and regularize noisy boundaries, but it cannot reliably recover a category, part, or instance that no source model ever exposed. The relevant question is therefore not only which 2D model is used, but what object it produces, at what granularity, with which identity and confidence, and when vocabulary enters the system.

\subsection{Masks provide support but not 3D identity}

Promptable and automatic mask generators are the dominant source of object support. SAM masks may be generated independently per image, prompted from an evolving 3D mask, or linked by a tracker. Independent masks maximize coverage but have arbitrary per-view IDs. SA3D closes a loop between a developing 3D target field and new SAM prompts, whereas SAM3D projects automatic masks and resolves identity in point space \citep{cen_2023_sa3d,yang_2023_sam3d}. Gaussian Grouping imports globally associated video-mask IDs before learning its identity field \citep{ye_2024_gaussian_grouping}. These pipelines use related mask technology but place the identity decision at different stages.

Mask topology also fixes a provisional scale. One automatic proposal may cover a mug, its handle, or a table setting. LangSplat samples whole, part, and subpart regions; OmniSeg3D learns hierarchy-aware affinity from nested masks; SAGA conditions Gaussian affinity on physical scale \citep{qin_2024_langsplat,ying_2023_omniseg3d,cen_2025_saga}. A method that consumes only one proposal level inherits a harder coverage ceiling than a carrier that retains overlapping or scale-conditioned evidence.

\subsection{Language can name pixels, regions, or entities}

Vision-language supervision enters at several spatial units. Dense models such as OpenSeg provide pixel-aligned language features used by OpenScene and OpenNeRF \citep{peng_2023_openscene,engelmann_2024_opennerf}. Crop-based systems encode an object proposal and its context: OpenMask3D ranks visible views and averages CLIP embeddings over several crop scales \citep{takmaz_2023_openmask3d}. Detectors and grounding models instead produce labeled boxes or text-conditioned masks. Open-YOLO 3D distributes YOLO-World box labels to 3D proposals, while InfoGS uses Grounding DINO to obtain the prompt mask that activates an already shaped carrier \citep{boudjoghra_2025_open_yolo3d,zhang_2025_infogs}.

Generated text expands supervision beyond class names. PLA creates captions at several spatial scales; RegionPLC uses regional descriptions; MPEC and PGOV3D insert entity descriptions or generated vocabularies between images and points \citep{ding_2023_pla,yang_2024_regionplc,wang_2025_mpec,zhang_2025_pgov3d}. These sources can describe attributes and affordances absent from fixed benchmarks, but they may hallucinate entities, omit visually small objects, or repeat correlated language biases. Text provenance and spatial support are therefore part of the supervision, not metadata that can be discarded after embedding.

\subsection{Self-supervised features contribute structure}

Language embeddings are semantically rich but often weak at boundaries and fine correspondence. Self-supervised image encoders such as DINO are commonly added as structural evidence. LERF uses DINO regularization alongside multi-scale CLIP; LEGaussians combines CLIP and DINO before quantization; HumanCrafter freezes DINOv2 and distills its features through reconstruction attention into a human semantic Gaussian field \citep{kerr_2023_lerf,shi_2024_legaussians,pan_2025_humancrafter}. The resulting 3D feature is not automatically open-vocabulary: HumanCrafter's final interface is a supervised 28-part classifier. The source representation and exposed query vocabulary must be reported separately.

Combining teachers also creates a competence-allocation problem. SAS integrates several 2D priors and estimates category-wise teacher capability rather than assuming every source is equally reliable \citep{li_2025_sas}. A teacher that is strong on large stuff classes may be poor on small instances or parts. Multi-teacher generation is useful only when the carrier retains enough source information or calibration to avoid averaging incompatible errors into a confident vector.

DINO features also support non-text semantic structure. SceneDINO predicts a continuous 3D DINO field from one image and evaluates it through unsupervised closed-class clustering, while DITR injects projected DINOv2 features into a supervised point network and distills an image-free variant \citep{jevtic_2025_scenedino,abouzeid_2026_ditr}. A foundation feature source does not itself determine the output vocabulary or evaluation task.

\subsection{Tracking, depth, and pose generate relational evidence}

A tracker contributes more than another mask: it asserts temporal identity. Gaussian Grouping, Segment then Splat, and InfoGS use tracked mask relationships to create persistent object sets, identity fields, or positive Gaussian pairs \citep{ye_2024_gaussian_grouping,lu_2025_segment_then_splat,zhang_2025_infogs}. The assertion can bridge weak appearance across views, but an identity switch becomes correlated supervision repeated throughout the scene. Geometry-based reconciliation can reduce dependence on tracking, as Gaga demonstrates with overlap between rendered Gaussian supports \citep{lyu_2026_gaga}.

Depth and pose define where evidence may be lifted. They can be measured, reconstructed, or predicted jointly with semantics. OpenScene and ConceptFusion assume calibrated RGB-D observations \citep{peng_2023_openscene,jatavallabhula_2023_conceptfusion}. Uni3R and Ov3R instead infer geometry from unposed images or video while also constructing semantic representations \citep{sun_2026_uni3r,gong_2026_ov3r}. Removing sensor inputs increases deployment flexibility but couples semantic error with learned correspondence: a wrong point map relocates otherwise correct 2D evidence.

Any3DIS makes tracking conditional on 3D-aware pivot selection and runs SAM2 in both directions before lifting a proposal; WildSeg3D instead precomputes a SAM2 mask cache over feed-forward point maps \citep{nguyen_2025_any3dis,guo_2025_wildseg3d}. The former can lock tracking error into proposal topology; the latter can reuse an alignment or cache error across every later prompt.

OV3DSeg-VGGT couples this relational evidence to a geometry foundation model: SAM2 tracks define cross-view instances, and OpenCLIP supplies one semantic target per instance before a VGGT head is adapted \citep{zhou_2026_ov3dseg_vggt}. The masks determine both positive-pair identity and achievable granularity. Unlike post-hoc lifting, a missing or over-merged mask changes the representation that will later be queried.

\subsection{Generation should expose uncertainty and absence}

Most pipelines retain only the selected mask, feature, or label. This makes missing evidence indistinguishable from confident background and makes systematic teacher error difficult to diagnose. Better generation records include proposal confidence, view and crop provenance, competing masks, tracker confidence, vocabulary source, and explicit absence. OpenNeRF uses cross-view feature variance to request new observations, VALA estimates visibility-aware feature reliability, and OGScene3D retains confidence for later map revision \citep{engelmann_2024_opennerf,wang_2025_vala,zhu_2026_ogscene3d}. These methods point toward a broader principle: generation should yield evidence with uncertainty and provenance, because association and fusion cannot reconstruct information that was collapsed before entering 3D.

\subsection{Generation can be offline, synthetic, or task-driven}

Large-scale part models make generation an offline data engine. Find3D combines grid-prompted SAM masks, Gemini part names and SigLIP embeddings before back-projecting pseudo-parts onto Objaverse shapes \citep{ma_2025_find3d}. PatchAlign3D reuses this corpus but additionally distills dense DINOv2 evidence into local point patches before language alignment \citep{hadgi_2026_patchalign3d}. The multi-view generator disappears at deployment, but its coverage and naming biases remain encoded in the reusable 3D model.

PartSLIP illustrates a deliberately coarse generation interface: GLIP returns text-conditioned boxes on ten rendered point-cloud views, and 3D voting converts their support into semantic and instance parts \citep{liu_2023_partslip}. COPS instead lifts dense DINOv2 features from 48 RGB-D renders, postponing language until after geometric clustering \citep{garosi_2025_cops}. Their contrast separates two questions often conflated as "using a VLM": whether 2D generates the decomposition itself and whether it only names an independently constructed part.

CoSMo3D moves generation further offline. An LLM helps organize a cross-category canonical training corpus, but deployment uses a Point Transformer and SigLIP text directly on the shape \citep{jin_2026_cosmo3d}. The 2D/language teacher is therefore absent at query time while its spatial assumptions remain embedded in the learned carrier. Such amortized systems should report data-engine provenance and canonicalization cost even when their measured inference is fast.

Rendered appearance need not imitate RGB. SAP renders surface normals and curvature from a Poisson-reconstructed point cloud, feeds their fused images to SAM2, and later uses CLIP only to name clustered geometric primitives \citep{bai_2025_sap}. This makes the 2D model sensitive to geometric cues that ordinary photographs may suppress, but it also inserts reconstruction and render-design assumptions before mask generation. Synthetic-view methods should state which channels the foundation model actually sees.

Generation can begin from an action rather than an object noun. Fun3DU asks a frozen LLM to infer the contextual object and the functional part implicit in an instruction, uses open-world detection and segmentation to choose views of the context, and then converts Molmo point predictions into SAM masks \citep{corsetti_2025_fun3du}. This expands the interface from category retrieval to functionality, but inserts a new causal failure upstream: a plausible yet wrong linguistic decomposition can make every subsequent 2D mask and 3D lift internally consistent around the wrong target.

The 2D source need not observe RGB at all. RangeSAM rasterizes LiDAR coordinates and sensor values into a spherical pseudo-image, replaces the RGB stem and square windows of a pretrained SAM2 Hiera encoder, and trains a closed-vocabulary range decoder \citep{kuhn_2026_rangesam}. Here the transferred evidence is model initialization and architectural prior rather than per-scene masks. Calling both RangeSAM and a frozen SAM lifting pipeline ``SAM-based'' would hide the decisive difference between supervised adaptation and zero-shot evidence generation.

\section{Associate: From Image Evidence to 3D Support}

Association determines which 3D elements a 2D prediction is allowed to supervise. It is therefore not a clerical projection step: it defines the support, uncertainty, and failure correlations of every later semantic or instance decision. Across the literature, association falls into three broad families---calibrated lifting, renderer-weighted lifting, and learned or temporary intermediaries---and methods that use the same 2D model can behave differently because they choose different association operators.

\begin{figure}[H]
\centering
\includegraphics[width=\textwidth]{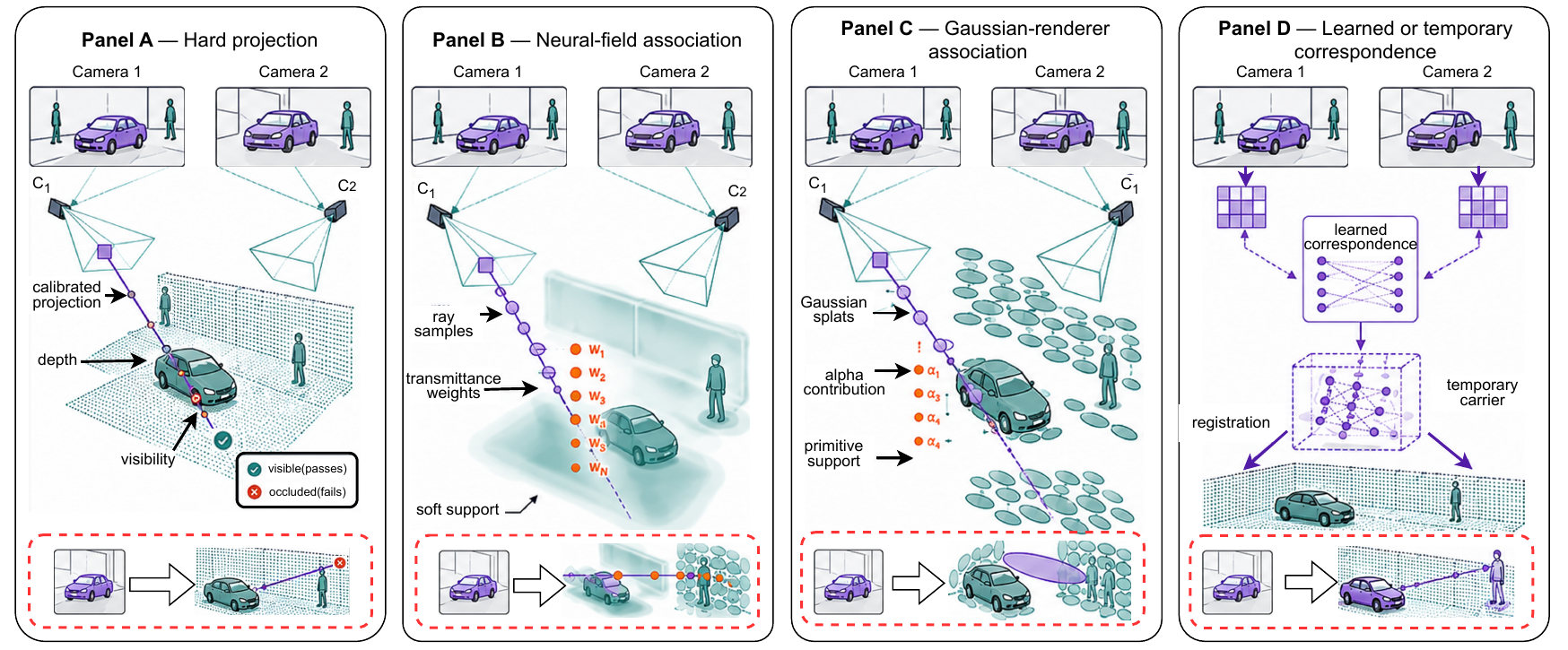}
\caption{Four realizations of 2D-to-3D association and their characteristic failure boundaries. Hard projection uses calibrated depth and an explicit visibility rule; neural fields distribute evidence through volumetric weights; Gaussian renderers use alpha-compositing contributions over explicit primitives; and learned or temporary correspondence introduces a predicted or reconstructed intermediary. Each mechanism locates evidence on candidate 3D support, but identity, semantic agreement, and granularity still require reconciliation.}
\label{fig:association-mechanisms}
\end{figure}

Figure~\ref{fig:association-mechanisms} separates four common realizations by splitting renderer-weighted association into neural-field and Gaussian-renderer cases. All four produce candidate 3D support; none, by itself, establishes whether evidence observed in different views belongs to the same physical entity.

\subsection{Calibrated lifting is hard correspondence with an explicit visibility policy}

For RGB-D sequences and registered point clouds, the most direct construction projects 3D points into each image using known intrinsics and extrinsics. A depth or visibility test then decides whether a projected point may inherit the pixel or mask evidence. OpenScene averages visible image features at each point before distilling them into a sparse 3D network, while ConceptFusion performs confidence-weighted online updates to a dense multimodal point map \citep{peng_2023_openscene,jatavallabhula_2023_conceptfusion}. Mask-driven systems use the same geometry for a different object: SAM3D and OV-SAM3D lift hard mask membership and then merge point sets, whereas Open3DIS uses mask-supported superpoints to create 3D proposals rather than merely naming existing ones \citep{yang_2023_sam3d,tai_2024_ov_sam3d,nguyen_2024_open3dis}.

The visibility rule is part of the method. A binary depth tolerance is simple but makes calibration noise a hard inclusion or exclusion. View ranking can instead select images in which an object proposal is well exposed, as in OpenMask3D \citep{takmaz_2023_openmask3d}. Details Matter makes this policy more explicit: it filters superpoints by both frame visibility and mask support, and computes tracking overlap only over co-visible superpoints \citep{jung_2025_details_matter}. These choices prevent an occluded surface from becoming negative evidence, but introduce thresholds that are tied to depth noise, superpoint size, and capture density. Consequently, two ``projection-based'' systems are not comparable unless their occlusion handling and 3D primitive are also specified.

\subsection{Renderers provide soft, visibility-aware association}

\subsubsection{Neural fields and Gaussian renderers}

Neural fields replace a binary point-pixel relation with a distribution along each camera ray. LERF, 3D-OVS, and OpenNeRF use NeRF density and transmittance weights to render language features; SA3D uses the same differentiable path in reverse to optimize a 3D mask field from prompted 2D masks \citep{kerr_2023_lerf,liu_2023_3d_ovs,engelmann_2024_opennerf,cen_2023_sa3d}. OpenNeRF additionally treats association as an acquisition policy: disagreement among projected OpenSeg features selects novel camera poses whose rendered RGB images are re-encoded. OmniSeg3D and GARField extend renderer association to affinity and scale-conditioned grouping \citep{ying_2023_omniseg3d,kim_2024_garfield}. The benefit is not only differentiability: a soft contribution weight represents partial visibility and shares evidence among samples on a ray. The cost is that semantic association inherits the renderer's geometric errors. Floaters, translucent density, or an inaccurate surface can distribute a correct 2D label to the wrong 3D support.

3D Gaussian Splatting supplies a related but discrete primitive. Gaussian Grouping, SAGA, LangSplat, Gaga, and OpenGaussian supervise attributes attached to splats through alpha-transmittance rendering \citep{ye_2024_gaussian_grouping,cen_2025_saga,qin_2024_langsplat,lyu_2026_gaga,wu_2024_opengaussian}. InfoGS makes the association nearly hard: each Gaussian inherits the 2D mask at the pixel to which it contributes most across all views, then those labels supervise correlations rather than stored class vectors \citep{zhang_2025_infogs}. FlashSplat instead accumulates each Gaussian's rendering contribution and derives an object membership vote without training a segmentation field \citep{shen_2024_flashsplat}. LEGaussians uses the same association path to optimize quantized CLIP-DINO features, with uncertainty controlling how strongly spatial smoothing should resolve view disagreement \citep{shi_2024_legaussians}. Thus, ``Gaussian lifting'' covers learned attributes, parameter correlations, and closed-form evidence accumulation; the renderer is shared, but the inference object is not.

CDSeg and SAGOnline show that renderer association can also terminate in explicit categorical labels. CDSeg records a binary pixel--primitive relation from projected support, front-to-back order, and transmittance-based visibility; the same operator labels either one-Gaussian-per-point carriers or native optimized scenes. SAGOnline adds a surface-crust visibility test and projected-center alignment before voting over SAM2-propagated masks from a rendered camera trajectory \citep{sun_2026_cdseg,sun_2025_sagonline}. Both avoid a learned scene-specific semantic field, but their hard support rules discard weak or misaligned evidence instead of representing association uncertainty.

Task-aware and domain-specific systems make view policy part of association. Fun3DU scores context masks and retains a default set of 50 views before lifting functional masks with multi-view agreement \citep{corsetti_2025_fun3du}. Wheat3DGS instead lifts one wheat-head mask into a Gaussian membership hypothesis, renders it into nearby cameras and searches for the best overlapping masks before refinement \citep{zhang_2025_wheat3dgs}. The first filters observations before correspondence; the second uses the developing 3D identity to seek correspondence. Both demonstrate that a projected pixel is only the start of association: view admission and cross-view search define which evidence is allowed to accumulate.

\subsubsection{View selection and mask graphs}

Mask-graph methods make the observation set explicit. SAI3D first over-segments geometry and converts SAM co-membership across visible views into a pairwise superpoint affinity \citep{yin_2024_sai3d}. MaskClustering instead keeps every CropFormer mask as a node and defines an edge by the fraction of other views that contain both projected mask supports \citep{yan_2024_maskclustering}. These are not equivalent overlap heuristics: SAI3D asks whether geometric primitives repeatedly share masks, whereas MaskClustering asks whether two masks receive common third-view support.

PS3 reverses the usual order for parts. It back-projects Semantic-SAM masks to create texture-aligned superpoints inside object proposals, then selects a viewpoint-dependent pivot and invokes SAM2 tracking to attach a masklet to each candidate \citep{yen_2026_ps3}. Association therefore determines the 3D primitive before reconciliation. This helps distinguish adjacent planar cabinet doors, but the chosen 2D granularity becomes a hard ceiling on the part inventory.

\subsubsection{Hybrid support operators}

Rethinking OVRF uses a hybrid path: renderer geometry defines samples, but supervision is point-wise before transfer to splats \citep{lee_2025_rethinking_ovrf}. DITR instead unprojects DINOv2 features into point-network skip connections \citep{abouzeid_2026_ditr}. The former risks cross-carrier mismatch; the latter risks calibration and visibility aliasing. Both show that "using 2D features in 3D" is underspecified without the support operator.

WildSeg3D uses MASt3R point maps as a temporary carrier for global alignment and cached mask groups \citep{guo_2025_wildseg3d}. ZeroPS renders planned views of an object cloud so that SAM and GLIP can provide part evidence before it is returned to points \citep{xue_2025_zerops}. View planning and reconstruction quality are therefore part of association, not neutral preprocessing.

Lifting by Gaussians hardens renderer association by assigning a mask to the per-pixel maximum-contributing Gaussian, then incrementally merging fragments with spatial and semantic evidence \citep{chacko_2025_lbg}. Trace3D preserves a richer view-by-instance weight matrix and uses Gaussian consistency to correct the masks themselves \citep{shen_2025_trace3d}. Hard dominant registration is simple and fast, while a multi-view matrix supports diagnosis and correction at higher construction cost.

LUDVIG exposes a third point on this spectrum: normalized renderer contribution weights form a closed-form multi-view feature aggregator, so arbitrary DINO, SAM or CLIP evidence can be attached to a pretrained 3DGS without feature optimization \citep{marrie_2025_ludvig}. This makes inverse rendering an explicit association operator; optional graph diffusion belongs to reconciliation and must be costed separately.

\subsection{Association may pass through a temporary carrier}

\subsubsection{Temporary geometric intermediaries}

Recent systems show that the representation performing association need not be the representation returned to the user. PointGS reconstructs a dense Gaussian intermediary, learns view-consistent affinities from rendered masks, clusters the Gaussians, and registers the result back to the original point cloud \citep{song_2026_pointgs}. This makes 3DGS a temporary visibility engine rather than the final semantic map. Conversely, Ov3R predicts point maps directly from video and fuses their descriptors after geometric registration, removing supplied camera poses but making reconstruction and association errors jointly learned \citep{gong_2026_ov3r}. PGOV3D lifts MLLM-generated image vocabularies and Grounded-SAM masks into partial point clouds, then uses inter-frame consistency before transferring the representation to full reconstructed scenes \citep{zhang_2025_pgov3d}.

HumanCrafter provides a domain-specific learned alternative: reconstruction attention aggregates frozen DINOv2 features, and predicted depth plus offsets unproject pixel-aligned tokens into a feed-forward human Gaussian field \citep{pan_2025_humancrafter}. These designs expose a useful audit question: where is correspondence first made, and where is the result finally stored? Direct projection, renderer weights, and learned registration make different assumptions about geometry and visibility. Reporting only the final carrier hides the stage at which an irreversible support error enters the pipeline. The next stage---reconciliation---cannot reliably repair an association that has already assigned evidence to the wrong physical surface; it can only decide how competing assignments should coexist. GeoSAM2 and MV3DIS show two ways for 3D geometry to guide association rather than merely receive it. GeoSAM2 renders normal and point maps, adapts SAM2 with geometric features, then visibility-votes prompted masks back to a shape \citep{deng_2026_geosam2}. MV3DIS projects coarse superpoint regions into images as common references for cross-view mask matching, while depth and visibility weights suppress occlusion errors \citep{zhao_2026_mv3dis}. The former prioritizes precise interaction; the latter prioritizes automatic scene-wide consistency.

PartSLIP and COPS expose the representation dependence of association. PartSLIP back-projects boxes, then votes their labels over superpoints; a generous box creates ambiguous support that later grouping must repair \citep{liu_2023_partslip}. COPS back-projects dense features instead and aggregates spatial neighbors plus feature-space neighbors before clustering \citep{garosi_2025_cops}. Both use known render cameras, yet the object associated to a point is either a discrete detection vote or a continuous descriptor. Projection accuracy alone does not make these evidence types interchangeable.

\subsubsection{Online and object-conditioned association targets}

Three online maps locate a further design split. ConceptGraphs back-projects each SAM region and compares it with persistent object nodes using both language-feature similarity and geometric overlap; Open-Fusion renders the current TSDF confidence channels and matches them to incoming SEEM regions; HOV-SG turns overlapping back-projected segments into a local graph before merging connected fragments \citep{gu_2024_conceptgraphs,yamazaki_2024_openfusion,werby_2024_hovsg}. The association target is therefore respectively an object, a semantic dictionary channel, or a segment hypothesis. The same posed RGB-D observation can produce different irreversible state because the receiving unit differs.

Object-conditioned systems acquire evidence selectively. OVI-MAP requests SigLIP semantics only from views that expand spherical surface coverage \citep{deng_2026_ovimap}. OpenFunGraph, FunGraph, OP3DSG and FunFact instead use object context to generate candidate parts before depth lifting \citep{zhang_2025_openfungraph,rotondi_2025_fungraph,kim_2026_op3dsg,fu_2026_funfact}. A handle is therefore not merely a smaller mask: its admissible support depends on a proposed parent, and parent errors become association errors.

TrackRef3D shifts association from isolated images to complete camera trajectories \citep{tan_2026_trackref3d}. Florence-2 and SAM2 produce noisy per-view discoveries, DEVA links them through time, and synonym clustering plus voting assigns one canonical identity before description generation. This improves cross-view consistency, but a track switch now contaminates every downstream mask, caption and Gaussian feature associated with that trajectory.

ProFuse uses one dense correspondence graph to establish both Gaussian geometry and cross-view mask identity \citep{chiou_2026_profuse}. Bidirectionally consistent warped overlap connects masks into 3D Context Proposals, while the same matches triangulate a compact reconstruction. This reuse is efficient, but it correlates failure: a bad match can seed the wrong surface and merge the wrong semantic evidence. Geometry and proposal association therefore require separate diagnostics even when computed from one primitive.

\subsubsection{Association shared with reconstruction}

Two systems sharpen this distinction. CoSegGaussians explicitly inverse-renders frozen DINO evidence onto reconstructed splats before a shared spatial decoder, so association precedes compression \citep{dou_2024_coseggaussians}. CCGS instead constructs DUSt3R pointmaps and uses their pixel correspondences to redefine inter-view SAM-mask overlap before partial Hungarian matching \citep{hu_2025_ccgs}. The first trusts geometry to gather continuous evidence; the second uses geometry to revise discrete identity correspondence. Both require the reconstruction and teacher passes in any honest cost ledger.

3DIML similarly separates association from field repair: global retrieval and dense keypoint matches create a mask graph, Leiden communities produce provisional identities, and only then does a label NeRF interpolate omissions \citep{tang_2024_3diml}. SAM2Object replaces image matching with bidirectional SAM2 tracks and filters them before forming superpoint affinities \citep{zhao_2025_sam2object}. These systems show that a stronger temporal or image correspondence front end changes which conflicts reach the 3D carrier; it does not eliminate the need to expose discarded masks and merge provenance.

\subsubsection{Query-specific association}

Interactive association can be query-specific. 100Editor back-projects clicks through rendered depth, reprojects them into arbitrary cameras as SAM2 prompts, and lifts the returned masks into a Gaussian region-of-interest set \citep{wu_2026_100editor}. This is an efficient interface for editing, but every view shares the initial depth error. A cross-view-consistent mask is not necessarily attached to the intended surface.

\section{Reconcile: Identity, Conflict, and Granularity}

Association produces observations; reconciliation decides whether observations refer to the same entity, which evidence should survive disagreement, and at what granularity an entity exists. These questions are coupled. A chair seat and a chair may be a split identity at object scale but a valid hierarchy at part scale, while two adjacent chairs may have nearly identical language features but must remain distinct instances. Reconciliation must therefore preserve geometry, identity, and semantic scale rather than treating feature averaging as a universal solution.

IQGS exposes a query-identity trade-off on 2DGS \citep{gao_2026_iqgs}. Fixing query indices to DEVA object IDs stabilizes within-scene binary-mask learning under view-dependent depth ordering, whereas Hungarian matching removes that binding and improves direct scene transfer at the cost of lower within-scene accuracy. Reconciliation is thus also a decision about which identities the carrier is allowed to memorize.

\subsection{Identity has an establishment time}

In externally reconciled systems, identity exists before the 3D carrier is learned. Gaussian Grouping consumes IDs from a video tracker, FlashSplat can accumulate masks with propagated IDs, and Segment then Splat uses SAM2 tracks to initialize object-specific Gaussian sets \citep{ye_2024_gaussian_grouping,shen_2024_flashsplat,lu_2025_segment_then_splat}. Their carriers can preserve identity efficiently, but cannot retroactively correct a tracker that merged two objects or changed identity across an occlusion.

Geometry-first methods delay identity until observations meet in 3D. SAM3D merges masks by bidirectional point overlap, OV-SAM3D updates a coarse mask-overlap table, and Gaga maintains a set-valued memory bank whose members are joined by cross-view overlap \citep{yang_2023_sam3d,tai_2024_ov_sam3d,lyu_2026_gaga}. Open3DIS hierarchically clusters superpoint-supported proposals using geometric overlap and feature similarity \citep{nguyen_2024_open3dis}. Details Matter strengthens this family with co-visible tracking, multi-view consensus pruning, iterative merging of partial proposals, and asymmetric removal of contained duplicates \citep{jung_2025_details_matter}. These operations improve proposal topology, yet their decisions are usually irreversible. An aggressive inclusion rule can remove a real small object before language classification ever sees it.

Learned identity carriers move reconciliation into optimization. OpenGaussian combines cross-view feature consistency with discrete coarse/fine codebooks, while InstanceGaussian and Identity-aware Language Gaussian Splatting learn instance-aware Gaussian attributes rather than relying only on post-hoc clustering \citep{wu_2024_opengaussian,li_2025_instancegaussian,jang_2025_identity_lgs}. PairGS persists pairwise grouping evidence, allowing identity to be represented by relations rather than a single categorical code \citep{cha_2026_pairgs}. InfoGS goes further from a categorical carrier: tracked SAM masks define positive and negative pairs, and contrastive activation shaping stores whether Gaussians should respond together under later parameter perturbations \citep{zhang_2025_infogs}. The resulting carriers can express object structure more richly, but they do not remove the need for an association prior: learned codes and correlations are still supervised through projected masks and renderer contributions.

CCL-LGS makes the boundary between identity and meaning especially clear \citep{tian_2025_ccl_lgs}. A zero-shot tracker first decides which SAM masks correspond across views; global and local CLIP evidence is then reconciled through a compact codebook whose pull and push objectives suppress view-dependent conflict. Better code separation cannot recover a missed or wrongly tracked object, while aggressive compression can merge a rare concept into a dominant code. Reconciliation therefore needs separate audits for correspondence error and semantic-code collapse.

Recent tracking pipelines make the establishment time explicit. Any3DIS reconciles forward/backward SAM2 tracks into class-agnostic proposals before optional CLIP or LLM naming, while PointSeg matches point and box prompts before iterative 2D refinement \citep{nguyen_2025_any3dis,he_2025_pointseg}. In both cases semantic naming cannot repair a missing or merged proposal, and PointSeg's workshop evidence should be weighted below archival main-conference results.

\subsection{Semantic conflict is not identical to identity conflict}

Multiple views can agree that a 3D support belongs to one object while disagreeing about its name, attributes, or description. RegionPLC explicitly reconciles redundant and conflicting regional captions before using them as point-language supervision \citep{yang_2024_regionplc}. MPEC instead inserts an entity layer: masked point-to-entity contrast establishes geometric consistency across views, and entity-to-language contrast aligns the pooled entity with text \citep{wang_2025_mpec}. This separation is valuable because background points and neighboring entities become explicit negatives. However, an incorrect entity mask corrupts both stages, and generated descriptions can turn a proposal error into a persistent semantic bias.

Contextual reasoning changes the conflict policy again. OV3D-CG selects a best view and constructs bounding-box, landmark, and mask representations, then asks an MLLM to reason about the instance in its scene \citep{zhou_2025_ov3d_cg}. Context can disambiguate visually similar objects, but can also encourage a room-prior shortcut: a plausible label may be unsupported by the object's pixels or geometry. Reconciliation with an MLLM therefore needs abstention and evidence attribution, not only a more capable classifier.

SOLE moves semantic reconciliation earlier still: projected CLIP features, generated captions, and softly matched noun entities supervise the 3D proposal decoder itself \citep{lee_2025_sole}. This can make mask topology responsive to free-form language, but also means linguistic ambiguity is no longer confined to post-hoc classification. COS3D instead preserves separate language and instance fields, maps instance features into language space during training, and sends text relevance back into the instance field at inference \citep{zhu_2025_cos3d}. The two designs expose opposite risk profiles: early semantic proposal learning can bake bias into every mask, while bidirectional query refinement can amplify a false relevance seed.

DiSCO-3D delays one form of reconciliation until query time: it discovers DINO-derived sub-concepts over a frozen LERF or OpenNeRF carrier and selects those relevant to the CLIP query \citep{petit_2025_disco3d}. ZeroPS instead reconciles SAM part masks and GLIP labels across planned views using its transfer modules \citep{xue_2025_zerops}. These examples show that conflict can be resolved over a fixed feature field or over lifted discrete parts, with different reversibility and failure modes.

Trace3D closes a feedback loop between reconciliation and geometry: mask disagreement defines ambiguous Gaussians, which are split and eventually pruned if ambiguity persists \citep{shen_2025_trace3d}. Reconciliation therefore mutates the reconstructed carrier and should be evaluated for both segmentation gain and appearance damage.

Semantic disagreement can now trigger reconstruction changes. COB-GS splits splats with high semantic-gradient ambiguity and repairs their texture, ObjectGS grows and prunes discrete-ID object anchors, and BEA-GS applies different losses to visible boundary splats and hidden splats exposed by extraction \citep{zhang_2025_cobgs,zhu_2025_objectgs,mazzucchelli_2026_beags}. These methods turn reconciliation into geometry governance, so appearance preservation and segmentation must be audited jointly.

NG-GS instead preserves the Gaussian mask carrier while interpolating evidence around variance-selected boundary splats through an RBF/hash-encoded NeRF module \citep{he_2026_nggs}. This makes reconciliation locally continuous without pretending that the discrete carrier itself has become continuous; topology and runtime therefore remain necessary evaluation dimensions.

Proto-SaGa reconciles identities before deployment by combining Gaussian depth with view-specific semantic probabilities in an instance memory, then replacing the disposable classifiers with cross-view feature prototypes \citep{oh_2026_proto_saga}. LUDVIG performs a learning-free alternative: graph edges multiply geometric proximity by DINOv2 similarity before diffusing coarse evidence \citep{marrie_2025_ludvig}. Both reduce raw mask inconsistency, but neither should collapse provenance into a single confident answer without exposing competing views or graph edges.

\subsection{Granularity must remain a query variable where possible}

\subsubsection{Scale and hierarchy as query variables}

Several carriers hard-code granularity. LangSplat uses whole/part/subpart levels, and tracker-based object sets usually inherit the mask generator's chosen scale \citep{qin_2024_langsplat}. Continuous-scale approaches instead condition affinity on physical scale: LERF selects a CLIP scale at query time, GARField uses scale-conditioned grouping, and SAGA learns a scale-gated Gaussian affinity \citep{kerr_2023_lerf,kim_2024_garfield,cen_2025_saga}. OmniSeg3D learns hierarchy-aware affinities from nested masks, while S2AM3D makes scale a prompt for point-cloud parts \citep{ying_2023_omniseg3d,su_2026_s2am3d}.

No formulation removes the coverage ceiling of its 2D evidence. A continuous scale cannot recover a semantic part absent from all masks, and a learned hierarchy may join spatially separate but semantically similar regions. The appropriate reconciliation output is therefore not always one final partition. Overlapping memberships, merge provenance, confidence, and a part-whole graph are often better persistent state, because they postpone a destructive granularity choice until the query is known.

Cross-shape part models add a second reconciliation axis: object pose. CoSMo3D uses semantic contrast, canonical map anchoring and box calibration to make analogous parts occupy a latent cross-category reference frame \citep{jin_2026_cosmo3d}. This replaces per-query multi-view voting with training-time spatial reconciliation, but turns canonical-frame construction into a source of systematic error. Rotation robustness should therefore be reported beside ordinary part mIoU, and canonical alternatives should remain auditable rather than being treated as neutral preprocessing.

DiscoNeRF makes identity competition explicit by rendering a finite set of categorical object slots and using per-view Hungarian matching against automatic SAM masks \citep{dumery_2025_viewconsistent}. False-positive and 3D total-variation losses then discourage overlapping or spatially noisy channels. This avoids clustering a high-dimensional feature field after training, but a fixed slot budget changes the failure mode: disconnected objects may share an identity even when each rendered boundary looks plausible.

DCSEG and OpenSplat3D retain a learned Gaussian feature followed by clustering, but decouple semantics at different strengths. DCSEG rasterizes HDBSCAN clusters and assigns them labels by matching or relaxed assignment to OpenSeg/OVSeg masks; a semantic teacher can be replaced without rebuilding the class-agnostic proposal field \citep{wiedmann_2025_dcseg}. OpenSplat3D also clusters SAM-contrastive Gaussian features, then attaches a multi-view MasQCLIP vector to each complete instance \citep{piekenbrinck_2025_opensplat3d}. In both, decomposition precedes naming, but the semantic unit differs: DCSEG resolves a framewise mask-assignment problem, whereas OpenSplat3D makes a persistent instance the language-retrieval atom.

Wheat3DGS offers a specialized hard-membership alternative. Its match-and-fine-tune loop alternates cross-view mask collection with binary optimization of which Gaussians belong to each wheat head \citep{zhang_2025_wheat3dgs}. The domain restriction supplies a strong object prior and enables downstream measurement, yet missed detector evidence and thin, occluded geometry become systematic selection bias rather than generic open-vocabulary error.

Before SAM2 trackers became common, SAI3D and MaskClustering demonstrated two scene-wide alternatives. SAI3D progressively lowers affinity thresholds and tests a candidate superpoint against multiple members of the growing region, limiting pairwise error chains \citep{yin_2024_sai3d}. MaskClustering filters under-segmented nodes and repeatedly merges connected components from high to low observer counts, recomputing consensus after each graph contraction \citep{yan_2024_maskclustering}. Both use global evidence, but one persists geometry-first regions and the other mask-first clusters.

PS3 adapts this reconciliation family to scene parts by comparing SAM2 masklets during breadth-first merging \citep{yen_2026_ps3}. Its MultiScan gains concentrate at AP50 rather than mean AP, and it misses tiny hooks on ScanNet++; this pattern is consistent with improved medium-scale identity but an unresolved fine-granularity coverage limit. A hierarchy-aware evaluation should therefore report both proposal topology and scale-specific recall.

Two hierarchy methods postpone a final partition in fundamentally different ways. N2F2 maps physical scale to nested prefixes of one feature vector, then combines all prefixes for a text query; granularity is implicit in the carrier rather than represented as parent--child nodes \citep{bhalgat_2024_n2f2}. Ultrametric Feature Fields instead learns a distance whose watershed threshold exposes different valid partitions, making hierarchy a stability property of the feature graph \citep{he_2024_ultrametric}. Neither mechanism guarantees semantic part identity: the former can conflate physical extent with linguistic detail, while the latter can let one bottleneck edge restructure an entire branch.

Search3D takes the explicit alternative: scene, object and part nodes are constructed separately and keep descriptors at their own level \citep{takmaz_2025_search3d}. This makes an object--part relation inspectable and lets material queries span multiple nodes, but it fixes the available levels to the proposal pipeline. Together these methods show that "hierarchical" may mean nested coordinates, threshold-stable partitions, or an explicit tree; evaluations should not treat those interfaces as interchangeable.

\subsection{Reconciliation extends to online, relational, and dynamic carriers}

ConceptGraphs and Open-Fusion show two forms of online reconciliation. The first performs greedy, thresholded semantic--geometric object assignment and later creates graph relations; the second solves an extended Hungarian assignment between rendered and observed confidence maps, rejecting low-soft-IoU matches \citep{gu_2024_conceptgraphs,yamazaki_2024_openfusion}. HOV-SG permits one incoming segment to connect several prior fragments before it builds floor and room parents \citep{werby_2024_hovsg}. Greedy object birth, one-to-one channel matching and graph-component merging encode different assumptions about whether fragmentation or collision is the more tolerable error.

Functional graphs add two reconciliation levels. FunGraph and OP3DSG merge part observations using parent, geometry, appearance and semantic evidence; FunFact then reconciles candidate functional edges with belief propagation over a dual factor graph \citep{rotondi_2025_fungraph,kim_2026_op3dsg,fu_2026_funfact}. CUA-O3D operates earlier, predicting teacher-specific uncertainty while distilling LSeg, DINOv2 and diffusion features \citep{li_2025_cua_o3d}. Node identity, relation identity and teacher agreement are distinct conflicts, even when all are called fusion.

Dynamic Gaussians reveal that a primitive's canonical index need not be its object identity. 4D LangSplat keeps tracked object supervision but represents semantics through a finite set of time-varying states; LangField4D learns an additional affiliation feature so identity can adapt when deformation moves a Gaussian across object support \citep{li_2025_4dlangsplat,xu_2026_langfield4d}. TIBR4D avoids a learned field and instead repeatedly recomputes instance membership after removing non-target Gaussians \citep{wu_2026_tibr4d}. These mechanisms respectively stabilize, adapt, or re-infer identity, and must be evaluated under occlusion and topology change rather than only smooth motion.

Earlier 4D methods expose two additional reconciliation choices. SA4D and SGG obtain persistent IDs from tracker-consistent masks, but SA4D maps them through a temporal field into a timestamped identity table whereas SGG attaches one code to a primitive with an explicit spacetime trajectory \citep{ji_2024_sa4d,wei_2026_sgg}. TRASE removes tracked labels and lets independent SAM masks shape a continuous feature space before DBSCAN assigns objects \citep{li_2026_trase}. Split4D also avoids global video IDs, but connects local mask evidence through moving primitives and temporally ordered optimization \citep{hu_2025_split4d}. Thus identity may be reconciled before lifting, after field learning, or implicitly through the reconstruction's motion support.

Contrastive Lift provides an earlier tracker-free field formulation: per-image Mask2Former or Detic identities supervise pairwise similarity, while a momentum-updated slow field supplies stable centers to the fast field \citep{bhalgat_2023_contrastive_lift}. It therefore removes both cross-frame tracks and a fixed slot budget, but does not remove reconciliation; correspondence is deferred into embedding dynamics and final clustering. CCGS takes the opposite route by explicitly reconciling pointmap-overlap masks before optimizing the Gaussian field \citep{hu_2025_ccgs}. These designs make the location of irreversibility visible: either a cluster becomes stable during learning or a discrete mask match becomes fixed before lifting.

Online query carriers add a lifecycle to reconciliation. EmbodiedSAM turns each SAM mask into a geometry-aware query and merges current and persistent queries by matrix similarity; ESAM++ keeps that logic while compressing the 3D encoder \citep{xu_2025_embodiedsam,liu_2026_esampp}. MoonSeg3R removes supplied depth and poses by combining explicit query indices with CUT3R's implicit recurrent state and a state-distribution token \citep{du_2026_moonseg3r}. The carrier can therefore be a query vector, a spatial index, or opaque reconstructive memory. All three need split, merge, disappearance and reappearance tests, not only final-scene AP.

\section{Fuse: Combining Evidence and Constructing the Carrier}

Fusion answers a different question from association: after evidence has been attached to 3D support and conflicts have been identified, how is repeated evidence converted into reusable state? The design space ranges from averaging to learned fields, discrete assignments, and relational graphs. A fusion rule should be judged by what uncertainty it preserves and which costs it moves from construction time to query time.

OV3DSeg-VGGT fuses semantics into a pretrained geometric head in two stages rather than optimizing a scene field \citep{zhou_2026_ov3dseg_vggt}. Cross-view instance contrast first makes DPT features discriminative; a projector then aligns them with CLIP, after which local KNN consensus smooths lifted point probabilities. This amortizes inference across scenes but couples semantic boundaries to VGGT depth quality.

\subsection{Averaging is a model, not a neutral default}

ConceptFusion performs confidence-weighted temporal averaging in an RGB-D point map, and OpenScene averages multi-view image features before selecting between distilled 2D and 3D predictions at query time \citep{jatavallabhula_2023_conceptfusion,peng_2023_openscene}. These rules are attractive because they are incremental and easy to audit. They are also unimodal summaries: mutually inconsistent observations collapse into one vector unless weights or uncertainty remain available. HumanCLIP similarly combines proposal occupancy with text similarity over points, which is efficient for repeated prompts after proposal extraction but inherits proposal coverage and threshold choices \citep{suzuki_2025_human_part}.

Robust aggregation changes what ``representative'' means. LEGaussians learns uncertainty-aware smoothing so that cross-view variance does not enforce uniform spatial consistency \citep{shi_2024_legaussians}. Octree-Graph's instance feature aggregation upweights views that are close to their own instance center and dissimilar from neighboring instances, making discriminability part of the fusion objective \citep{wang_2025_octree_graph}. VALA and PanoGS likewise treat view reliability and panoramic consistency as estimation problems rather than unweighted feature pooling \citep{wang_2025_vala,zhai_2025_panogs}. These methods suggest that a carrier should retain either uncertainty or the provenance needed to recompute it; a final mean alone cannot distinguish consensus from cancellation.

Learning-free fusion does not remove modeling choices. LUDVIG normalizes renderer support before aggregating features and Proto-SaGa averages masked features first within views and then across view-specific prototypes \citep{marrie_2025_ludvig,oh_2026_proto_saga}. The former assumes renderer contribution is sufficient reliability; the latter assumes prototype consensus suppresses association errors. Their low deployment complexity therefore depends on construction-time weighting choices that should remain auditable.

Amortized fusion moves a different set of choices into cross-scene training. PartField distills overlapping SAM2 and 3D part proposals into a continuous feature field whose clustering exposes multiple part granularities \citep{liu_2025_partfield}. This avoids repeated view rendering and segmentation at deployment, but the apparently cheap forward pass is backed by two weeks of training on eight A100 GPUs. Cross-shape consistency is therefore a learned carrier property, not evidence that hierarchy was recovered without supervision or cost.

\subsection{Learned fields trade query efficiency for optimization and compression choices}

Neural fields jointly fuse repeated supervision through a scene-specific continuous function. LERF, 3D-OVS, OpenNeRF, OmniSeg3D, and GARField differ in whether the field stores language relevance, localized pixel-aligned features, a label-conditioned semantic representation, scale-free affinity, or physical-scale-conditioned grouping \citep{kerr_2023_lerf,liu_2023_3d_ovs,engelmann_2024_opennerf,ying_2023_omniseg3d,kim_2024_garfield}. OpenNeRF also adds evidence after locating high-variance regions, so its fused field depends on an active observation loop. The shared advantage is smooth, renderer-consistent query output. The shared risk is entanglement: geometry errors and semantic errors are optimized through the same rendering path, and interpolation can blur an instance boundary that was crisp in 2D.

Gaussian carriers expose the compression choice more directly. LangSplat decodes a low-dimensional scene-specific latent into CLIP space; LEGaussians uses vector quantization; Dr. Splat registers dominant Gaussians and compresses a reusable feature index with product quantization \citep{qin_2024_langsplat,shi_2024_legaussians,kim_2025_dr_splat}. SuperGSeg moves language from individual splats to sparse Super-Gaussians, while ExtrinSplat keeps semantics in an external group-to-text index rather than inside every Gaussian \citep{liang_2026_supergseg,ding_2026_extrinsplat}. InfoGS adds another semantic primitive: a correlation pattern in the Gaussian attribute decoder rather than an explicit feature attached to each splat \citep{zhang_2025_infogs}. Its activation-level objective makes same-object Gaussians respond coherently after repeated parameter changes, supporting segmentation and editing while tying query semantics to the intervention used to expose the correlations. Compression and carrier placement are therefore not only engineering parameters; they constrain which overlaps, relations, and query operations can be represented.

Rethinking OVRF demonstrates cross-carrier fusion: direct point-wise semantic supervision first creates a NeRF language field, which is then transferred into 3DGS for faster rendering \citep{lee_2025_rethinking_ovrf}. SceneDINO instead amortizes joint density and DINO-feature prediction across scenes \citep{jevtic_2025_scenedino}. These routes move cost and error differently: transfer can mismatch supports, while amortization can smooth away scene-specific or occluded detail.

FMGS illustrates hybrid fusion: multi-scale CLIP supplies language while DINO sharpens feature alignment inside a hash field rendered by Gaussians \citep{zuo_2025_fmgs}. ObjectGS takes the opposite semantic primitive, binding one-hot IDs to object anchors so alpha blending cannot average identities as continuous vectors \citep{zhu_2025_objectgs}. The tradeoff is between dense open-text flexibility and a cleaner but scene-fixed discrete inventory.

\subsection{Discrete fusion preserves topology but can hide uncertainty}

SAM3D and OV-SAM3D produce hard point partitions through overlap-based merging, while FlashSplat derives Gaussian membership through an accumulated contribution vote \citep{yang_2023_sam3d,tai_2024_ov_sam3d,shen_2024_flashsplat}. Their output is directly usable for extraction and editing, and avoids decoding a dense semantic field for each query. Yet a hard ID suppresses alternative correspondences. Details Matter shows why this matters: proposal topology changes substantially after overlap removal, consensus refinement, iterative merge, and inclusion filtering, even before the classifier is changed \citep{jung_2025_details_matter}. A robust discrete carrier should therefore store the evidence behind an assignment or permit local revision.

CDSeg accumulates per-label votes and applies a fixed local majority filter without filling wholly unobserved support, whereas SAGOnline first creates sparse majority-voted Gaussian labels and then uses a learned image-space refinement network to densify rendered masks \citep{sun_2026_cdseg,sun_2025_sagonline}. The distinction is consequential: local 3D filtering changes the persistent carrier, while image-space refinement can improve a displayed mask without proving that the underlying native-3D labels are complete.

Dataset-trained models amortize fusion across scenes. PLA and RegionPLC learn from generated language at view, entity, and regional scales; MPEC uses point-entity-language contrast; PGOV3D transfers dense partial-view supervision to complete scans \citep{ding_2023_pla,yang_2024_regionplc,wang_2025_mpec,zhang_2025_pgov3d}. HumanCrafter similarly amortizes a frozen DINOv2-to-Gaussian semantic field, but fixes the output to 28 human-part classes and couples it to feed-forward reconstruction \citep{pan_2025_humancrafter}. This eliminates per-scene language-field optimization at deployment, but does not eliminate fusion cost: it relocates it into large-scale data generation and training. ``Training-free'' and ``annotation-free'' must consequently be qualified by whether reconstruction, 2D foundation-model inference, pseudo-label creation, and dataset training are counted.

SAM4D extends fusion into time and modality: image and LiDAR encoders share 3D positional coordinates, while motion-aware memory attention propagates prompts with ego-motion compensation \citep{xu_2025_sam4d}. Its state should be audited for cross-modal contamination because an erroneous temporal association can alter both output streams.

\subsection{Fusion should be evaluated as a full ledger}

A credible efficiency claim separates reconstruction, 2D preprocessing, association/reconciliation, carrier optimization, storage, and per-query inference. LEGaussians, for example, reports fast rendering after approximately 30 minutes of dense feature extraction and about one hour of semantic fitting in its setup \citep{shi_2024_legaussians}. RegionPLC reports low direct inference latency only after multi-GPU dataset training and regional-caption preprocessing \citep{yang_2024_regionplc}. Octree-Graph has a compact adaptive occupancy structure, but its map still follows segmentation, captioning, projection, and instance merging \citep{wang_2025_octree_graph}. The carrier is the result of the whole pipeline; evaluating only its last operation systematically favors methods that move work upstream.

Open-YOLO 3D provides a complementary efficiency point: it avoids SAM and dense CLIP maps by projecting low-granularity detector labels onto pretrained 3D proposals and voting across the most visible views \citep{boudjoghra_2025_open_yolo3d}. This is fast for a known prompt set, but a static scene with many sequential new prompts can favor a cached CLIP carrier. Efficiency is consequently workload-dependent, not a single scalar property. What is fused can also be a relation rather than a feature. CMAT converts back-projected DINOv3 descriptors into a pairwise affinity matrix and trains a point backbone to reproduce that organization before affordance-specific prompting \citep{huang_2026_affordance}. This avoids forcing exact cross-modal feature equality, but it can still transfer an appearance-driven relation when the desired distinction is functional.

SAP likewise fuses relations, but without training a backbone: multi-view SAM2 agreement forms a semantic distance between point-graph nodes, while coordinates, normals and curvature supply geometric distances \citep{bai_2025_sap}. Graph cutting and adaptive merging produce primitive instances before CLIP voting assigns types. The formulation is useful precisely because it does not collapse all cues into one descriptor, although its thresholds and Poisson geometry become part of the effective segmentation model.

Multimodal domain adaptation makes reliability an explicit fusion variable. Spoecklberger et al. combine frozen AM-RADIO image features, a trainable sparse LiDAR stream and an MLP fusion stream, then regularize the fused representation toward RGB for daylight targets and toward LiDAR at night \citep{spoecklberger_2025_modality}. This improves the reported average across four adaptation directions, but the predefined global preference becomes part of the deployment contract. Ambiguous local lighting, reflections or sparse geometry motivate per-point reliability rather than one environmental switch for a whole target domain.

OVIR-3D is a useful historical counterexample to purely offline fusion. It associates each incoming Detic region with persistent point instances, periodically filters and merges them using visibility, and preserves multiple clustered descriptors for later retrieval \citep{lu_2023_ovir3d}. Its reported near-real-time fusion therefore includes identity maintenance but not the complete reconstruction and 2D-proposal pipeline. More importantly, the online setting makes pruning irreversible: an early tiny-object deletion or large-object split can outlive the observation that caused it.

\subsection{Compact and temporal carriers make information loss explicit}

Fusion statistics also reveal what disagreement each carrier can preserve. ConceptGraphs count-averages one descriptor per object, Open-Fusion averages confidence maps within a matched dictionary entry, and HOV-SG uses DBSCAN to select a dominant feature from the point-wise observations of a merged segment \citep{gu_2024_conceptgraphs,yamazaki_2024_openfusion,werby_2024_hovsg}. Mean updates are cheap but can hide multimodality; majority clustering rejects outliers but may erase rare, correct views. Neither rule is a calibrated substitute for retaining contradictory evidence and assignment provenance.

Recent compact Gaussian carriers make the compression statistic explicit. LightSplat commits each Gaussian to a two-byte mask or cluster ID and keeps language in an external lookup table \citep{bang_2026_lightsplat}. SCOUP instead learns sparse code coefficients on 2D SAM/OpenCLIP regions, accumulates them through renderer weights and retains several dominant atoms per Gaussian \citep{budimir_2026_scoup}. A foreign key is cheaper and more interpretable; a sparse mixture can preserve polysemy and cross-view disagreement. Both are irreversible when filtering or Top-K selection removes the evidence required by a later query.

CoSegGaussians supplies a third compression statistic: DINO and spatial evidence are decoded by one shallow network shared across all Gaussians rather than stored as an independent dense vector on every primitive \citep{dou_2024_coseggaussians}. This reduces learnable storage, yet correlated decoder smoothing can turn a local boundary error into a scene-wide inductive bias. CCGS moves compactness into carrier geometry through same-class piecewise-plane constraints \citep{hu_2025_ccgs}. Parameter compactness and geometric compactness are therefore distinct claims and should be reported separately.

Dynamic fields make this distinction temporal. 4D-Editor stores PCA-compressed DINO evidence in separate static and time-conditioned semantic fields, then induces a target from one-frame strokes with recursive feature matching \citep{jiang_2023_4d_editor}. Unlike a persistent track ID, this carrier supports new object selections after construction; unlike an open-vocabulary field, it relies on visual similarity and hand-set distance scales. Query flexibility thus depends on both what the carrier stores and what evidence the query is allowed to introduce.

TEXTRIX moves fusion out of scene optimization altogether: a sparse DiT generates a native 3D attribute grid from a position map and a DINOv3-conditioned input view, using the same carrier for texture or part labels \citep{zeng_2026_textrix}. This is not multi-view evidence accumulation but amortized attribute generation. Its efficiency and boundary claims belong to a mesh-generation protocol, while its pseudo-label construction remains part of the supervision ledger.

\subsection{Fusion can preserve relational structure}

GeoCGA changes the fused object from a descriptor to a relation structure. It aligns an LLM-expanded semantic-spatial graph with an object-level graph over Gaussian clusters, then regularizes view-consistent Gaussian responses \citep{tao_2026_geocga}. Relational reasoning can resolve spatial queries that unary CLIP similarity cannot, but it inherits every missed object and incorrect edge from the proposal graph.

\section{Persist, Render, and Query}

Persistence determines what a system can answer without rerunning its 2D foundation models. A per-point label map, a continuous field, a Gaussian attribute, an object set, and a relation graph may all yield a colored segmentation, but they support different operations and failure recovery. The central question is not only where semantics are stored; it is which geometric, identity, uncertainty, and relational state remains available when a new query arrives. Table 3 compares the natural output, main advantage, and correction boundary of the principal carrier families. The comparison prevents a shared visualization---for example, a colored mask---from being mistaken for a shared query contract.

\CarrierComparisonTable

DiLEGS shows that persistent language Gaussians need not inherit a dense photometric reconstruction: synchronized SAM/SAM2 identities and CLIP features can be trained directly on a sparse SfM seed \citep{li_2026_dilegs}. The result is reconstruction-free only with respect to dense appearance optimization, since calibration, SfM support and mask preprocessing remain part of construction. FreeArtGS gives persistence a kinematic meaning: AllTracker correspondences and DINOv3 features initialize a two-part partition jointly optimized with articulated Gaussian geometry and joint state \citep{dai_2026_freeartgs}.

\subsection{Persistent carriers expose different query surfaces}

Point carriers support direct native-3D extraction. OpenScene and learned point-language models such as RegionPLC and MPEC compare stored point features with arbitrary text embeddings \citep{peng_2023_openscene,yang_2024_regionplc,wang_2025_mpec}. Hard point masks from SAM3D or OV-SAM3D support object extraction with little query-time computation, but their vocabulary and partition are largely fixed by upstream masks and tags \citep{yang_2023_sam3d,tai_2024_ov_sam3d}. PGOV3D stores a cross-scene point encoder rather than scene-specific semantic vectors; its reuse is model-level, with each new scan receiving features after inference \citep{zhang_2025_pgov3d}. These are all point-based systems, yet one persists scene state, another persists discrete identities, and another persists transferable parameters.

Neural fields persist a rendering function. LERF can render language relevancy from novel views and search over physical scale, OpenNeRF stores localized OpenSeg features and can acquire additional observations at construction time, whereas SA3D persists a prompted target mask through a radiance-field carrier \citep{kerr_2023_lerf,engelmann_2024_opennerf,cen_2023_sa3d}. This makes view synthesis and continuous spatial queries natural, but native 3D extraction depends on sampling the field and choosing thresholds. Gaussian carriers combine explicit primitives with differentiable rendering. SAGA supports scale-conditioned selection, OpenGaussian provides instance-level codes and CLIP features, and FlashSplat stores hard or overlapping object membership suitable for direct Gaussian editing \citep{cen_2025_saga,wu_2024_opengaussian,shen_2024_flashsplat}. InfoGS instead persists joint response under parameter perturbation, while HumanCrafter predicts a closed-set human-part feature field in one feed-forward pass \citep{zhang_2025_infogs,pan_2025_humancrafter}. These carriers may share Gaussian geometry while exposing fundamentally different query operations.

CDSeg makes this output contract explicit through two modes: an index-preserving point-completed carrier can return labels to the supplied points, while a native-scene carrier retains labels on optimized Gaussians for later rendering. SAGOnline instead persists a sparse discrete Gaussian label set whose rendered masks are refined online \citep{sun_2026_cdseg,sun_2025_sagonline}. Neither carrier stores a universal language feature; a new semantic query therefore depends on new upstream masks or detections rather than text lookup against the existing state.

Feed-forward reconstruction can also make the labeled 3D output itself the carrier. PanSt3R combines frozen DINOv2 semantics with MUSt3R geometry to produce globally consistent classes and instance IDs over unposed views, then optionally converts that point reconstruction to a panoptic 3DGS \citep{zust_2025_panst3r}. The distinction matters: its direct approximately 2.3-minute result and its approximately 35-minute LUDVIG novel-view variant answer different interface and cost questions.

COS3D demonstrates that one geometric carrier may persist two query surfaces: a compact discriminative instance field and a high-dimensional language field connected through an instance-to-language mapping \citep{zhu_2025_cos3d}. Query-time language-to-instance refinement then uses the former to repair the latter's boundaries. Persistence can therefore include an interaction rule between fields, not merely a collection of stored vectors.

Object sets and graphs expose a higher-level interface. Segment then Splat persists object-specific Gaussian sets across time, while ReLaGS and OGScene3D attach relational or confidence-aware graph state to scene entities \citep{lu_2025_segment_then_splat,xie_2026_relags,zhu_2026_ogscene3d}. Octree-Graph adds adaptive occupancy to each object node and spatial/semantic relations to edges, enabling text retrieval and path planning without returning to a dense point cloud for every operation \citep{wang_2025_octree_graph}. The graph is more expressive than unary segmentation, but its errors have wider blast radius: one incorrect merge can change occupancy, captions, edges, retrieval, and navigation together.

Persistence can also be a precomputed interaction cache. WildSeg3D stores aligned point maps and SAM2-derived multi-view groups, reducing later point-prompt interaction to milliseconds only after a reported tens-of-seconds construction stage \citep{guo_2025_wildseg3d}. SAM4D persists motion-aware camera-LiDAR memory for streaming prompts \citep{xu_2025_sam4d}. These carriers optimize different workloads: repeated interaction with a static reconstruction versus temporally evolving multimodal perception.

RelationField generalizes persistence from unary features to pairwise spatial queries. Its relationship response depends on a query location and a second field location, allowing subject-predicate-object evidence distilled from an MLLM to guide instance localization \citep{koch_2025_relationfield}. This differs from segmenting objects first and attaching graph edges afterward: the relation is part of the rendered field.

\subsection{Open vocabulary has a timing}

``Open-vocabulary'' does not specify when text can change. In 3D-OVS, the label set is supplied before per-scene training, so a new vocabulary can require retraining \citep{liu_2023_3d_ovs}. In CLIP-feature fields such as LERF or LEGaussians, language-aligned features are stored once and new text can be compared after construction \citep{kerr_2023_lerf,shi_2024_legaussians}. In proposal-first systems such as OpenMask3D, the instance geometry is built independently and text names it afterward \citep{takmaz_2023_openmask3d}. OV3D-CG postpones naming further, using an MLLM at query/classification time to incorporate descriptions and context \citep{zhou_2025_ov3d_cg}. Later vocabulary binding improves flexibility but increases per-query cost and cannot recover an object missing from the persistent proposal set.

The same distinction applies to generated-language supervision. PLA, RegionPLC, MPEC, and PGOV3D use open-ended text to train a reusable 3D encoder \citep{ding_2023_pla,yang_2024_regionplc,wang_2025_mpec,zhang_2025_pgov3d}. Their final classifier can accept new class names, but the representation has already been shaped by the entities, captions, and categories produced during training. Open text at inference does not imply open perceptual coverage.

REALM binds language at the latest end of this spectrum. It first stores a SAM-derived feature field on a reconstructed 3DGS, then lets an MLLM interpret an implicit expression, choose global and local views, and refine the selected identity in the field \citep{shi_2026_realm}. This supports compositional queries without rebuilding geometry, but it does not make each query a cheap vector lookup: the agent calls the MLLM twice and performs target-specific optimization. Query expressiveness and query-time state mutation should therefore be reported together.

\subsection{Rendering and native-3D extraction are different evaluation domains}

A rendered relevancy map evaluates the carrier through its renderer and current viewpoint. A native 3D mask evaluates support on points, mesh faces, voxels, or Gaussians, including surfaces that may be occluded in the image. These outputs should not be ranked in one table without a conversion protocol. LEGaussians' annotated novel-view results, Open3DIS' ScanNet200 instance AP, and RegionPLC's foreground semantic mIoU answer different questions even when all are described as open-vocabulary segmentation \citep{shi_2024_legaussians,nguyen_2024_open3dis,yang_2024_regionplc}. Similarly, path success from Octree-Graph tests occupancy and planning in addition to semantics \citep{wang_2025_octree_graph}.

The carrier should therefore be evaluated against its promised interface: novel-view localization for rendered fields, native 3D topology for masks, repeated-query latency and storage for feature carriers, update consistency for online maps, and edge/retrieval/navigation accuracy for graphs. Persistence is not a cosmetic implementation detail. It defines what is reusable, what can be revised, and which downstream claims the evidence can actually support.

\subsection{Hierarchies and graphs expose structured query interfaces}

The hierarchy-bearing carriers sharpen this interface distinction. N2F2 returns a composite rendered relevance field, Ultrametric Feature Fields returns a partition selected by a distance threshold, Search3D returns stored object/part supports from an explicit tree, and OVIR-3D ranks online-built instance segments \citep{bhalgat_2024_n2f2,he_2024_ultrametric,takmaz_2025_search3d,lu_2023_ovir3d}. All accept flexible downstream choices, but the adjustable variable is respectively text relevance, granularity threshold, hierarchy node, or language-ranked identity. "Queryable 3D representation" is therefore too broad for evaluation unless the returned support and mutable state are named.

ConceptGraphs, Open-Fusion and HOV-SG make that interface contrast concrete \citep{gu_2024_conceptgraphs,yamazaki_2024_openfusion,werby_2024_hovsg}. ConceptGraphs returns reusable object nodes whose captions and relations support retrieval and planning; Open-Fusion returns a text-selected TSDF surface or voxel support through a region dictionary; HOV-SG first narrows a query through floor and room parents and then selects an object and navigation goal. These carriers should be tested through object-update consistency, live semantic throughput and parent-aware retrieval respectively, not only a shared semantic mIoU.

KeySG makes retained images part of the persistent API: selected keyframes and hierarchical descriptions are retrieved so relationships can be inferred on demand instead of stored as fixed edges \citep{werby_2026_keysg}. OpenFunGraph and OP3DSG persist functional elements and relations, while FunFact additionally persists marginal confidence \citep{zhang_2025_openfungraph,kim_2026_op3dsg,fu_2026_funfact}. Query-time flexibility therefore depends on whether the carrier retains raw visual evidence, fixed symbolic edges, or a revisable probabilistic graph.

KNA-SG goes further by linking every object node to ID-marked keyframes and writing verified query relations back as memory \citep{xu_2026_knasg}. LOST-3DSG chooses the opposite compression point, storing compact label, color, material and description attributes for dynamic lifecycle updates \citep{ferraina_2026_lost3dsg}. OpenCity3D retains neither object graph: it projects a hierarchy of SigLIP features onto urban points for prompt-driven heatmaps \citep{bieri_2025_opencity3d}. Provenance-rich, compressed-symbolic and dense-field persistence support different correction operations.

DovSG and Open-World 3DSG add two persistence boundaries. DovSG locally invalidates voxels and rebuilds affected object nodes after relocalizing a changed RGB-D region \citep{yan_2025_dovsg}. Open-World 3DSG copies object descriptions, best views and relations into a vector index for later retrieval \citep{yu_2026_ow3dsg}. The former requires stale-evidence deletion; the latter additionally requires graph--index synchronization after any correction.

OpenVoxel offers a third choice: renderer-weighted SAM2 votes persist discrete voxel-group IDs, while DAM/Qwen3-VL captions form a readable scene map queried by text-to-text reasoning \citep{huang_2026_openvoxel}. This removes dense language-field training but not the information bottleneck; a distinction absent from the canonical caption or merged group cannot be reconstructed by a later prompt.

\subsection{Query-time and dynamic carriers change the reuse contract}

ReferSplat and GaussDet expose two different referring interfaces \citep{he_2025_refersplat,hassan_2026_gaussdet}. ReferSplat stores expression-aware Gaussian features learned from scene-specific Grounded-SAM pseudo-masks and position-aware language interaction. GaussDet instead renders persistent instance groups and aggregates discrete Qwen-VL or YOLO-World detections into a label distribution, allowing new queries without storing a dense CLIP field. The former adapts the carrier to the expression set; the latter freezes the carrier but inherits the detector and group topology. Their referring scores are therefore not interchangeable evidence of the same generalization regime.

ZeroSplat removes the semantic carrier altogether \citep{ding_2026_zerosplat}. For each free-form query it parses semantic labels and boxes, runs SAM3, lifts the surviving masks to Gaussians and applies spatial refinement. This supports zero, one or many targets without scene optimization or extra per-Gaussian features, but prevents cheap repeated lookup: semantic work is recomputed at query time rather than amortized into persistent storage.

Dynamic carriers must answer both where and when. 4D LangSplat mixes a small set of state prototypes; LangField4D represents continuous space-time semantics in a TetraPlane; ST4R-Splat first grounds a persistent instance and then queries its temporal state cache \citep{li_2025_4dlangsplat,xu_2026_langfield4d,meng_2026_st4rsplat}. 4D Synchronized Fields goes further by conditioning language on object-level kinematics learned with reconstruction \citep{barhdadi_2026_4dsync}. These interfaces return different objects: a per-frame mask, a continuous state score, or an object--time interval. Their metrics cannot be collapsed into static segmentation mIoU.

These dynamic systems further show that storage semantics precede interface semantics. DGD retains the full DINOv2 or CLIP teacher space on deforming Gaussians, enabling click or text similarity at high memory cost \citep{labe_2024_dgd}. 4-LEGS instead stores decoded ViCLIP evidence and queries an action jointly in space and time \citep{fiebelman_2025_4legs}. Multi4D compresses in the opposite direction: it discards transient appearance primitives from semantic learning, then stores a 32-dimensional field only on persistent geometry and maps a Grounding-DINO/SAM query to a cluster \citep{wang_2026_multi4d}. Open vocabulary therefore does not determine one carrier form; it can be a teacher vector, a video-language field, or an object index over persistent motion.

Online methods also clarify that persistence is an update rule, not merely a data type. EmbodiedSAM and ESAM++ maintain query identities while streaming RGB-D frames, whereas MoonSeg3R couples sparse query indices to a reconstructive model's latent state \citep{xu_2025_embodiedsam,liu_2026_esampp,du_2026_moonseg3r}. A bounded query bank is compact and immediately actionable, but a mistaken merge changes future evidence admission. Online benchmarks should therefore report identity transitions and recovery, not only the final accumulated mask.

Query-time carriers occupy the other extreme. 100Editor creates a Gaussian membership mask only after clicks are supplied, while TEXTRIX generates a complete native attribute grid from a conditioning view \citep{wu_2026_100editor,zeng_2026_textrix}. Both avoid storing a universal scene-language vector, yet their reuse contracts differ: one answers new spatial prompts on an existing reconstruction; the other amortizes a learned object-part prior over new meshes.

\section{Evaluation Protocols and the Full Cost Ledger}

Reported accuracy is comparable only when the prediction object, vocabulary, source geometry, and query protocol match. Published systems often use the same metric name for materially different tasks: mIoU may evaluate rendered pixels, native point labels, or explicit Gaussians; AP may evaluate oracle masks, proposals from a trained 3D network, proposals lifted from 2D masks, or a union of both. The LAF output contract therefore keeps protocol families separate rather than placing every number in one leaderboard.

IQGS adds rendered-view class-agnostic instance mIoU and a no-finetuning scene-transfer test, while OV3DSeg-VGGT evaluates semantic probabilities on predicted feed-forward geometry \citep{gao_2026_iqgs,zhou_2026_ov3dseg_vggt}. Neither denominator is interchangeable with open-vocabulary AP on fixed ScanNet geometry. IQGS reports 30k A100 iterations without normalized wall time; for OV3DSeg-VGGT, we withhold numerical cost and accuracy claims because only the publisher HTML, not a valid PDF, was locally verifiable.

\begin{figure}[H]
\centering
\includegraphics[width=\textwidth]{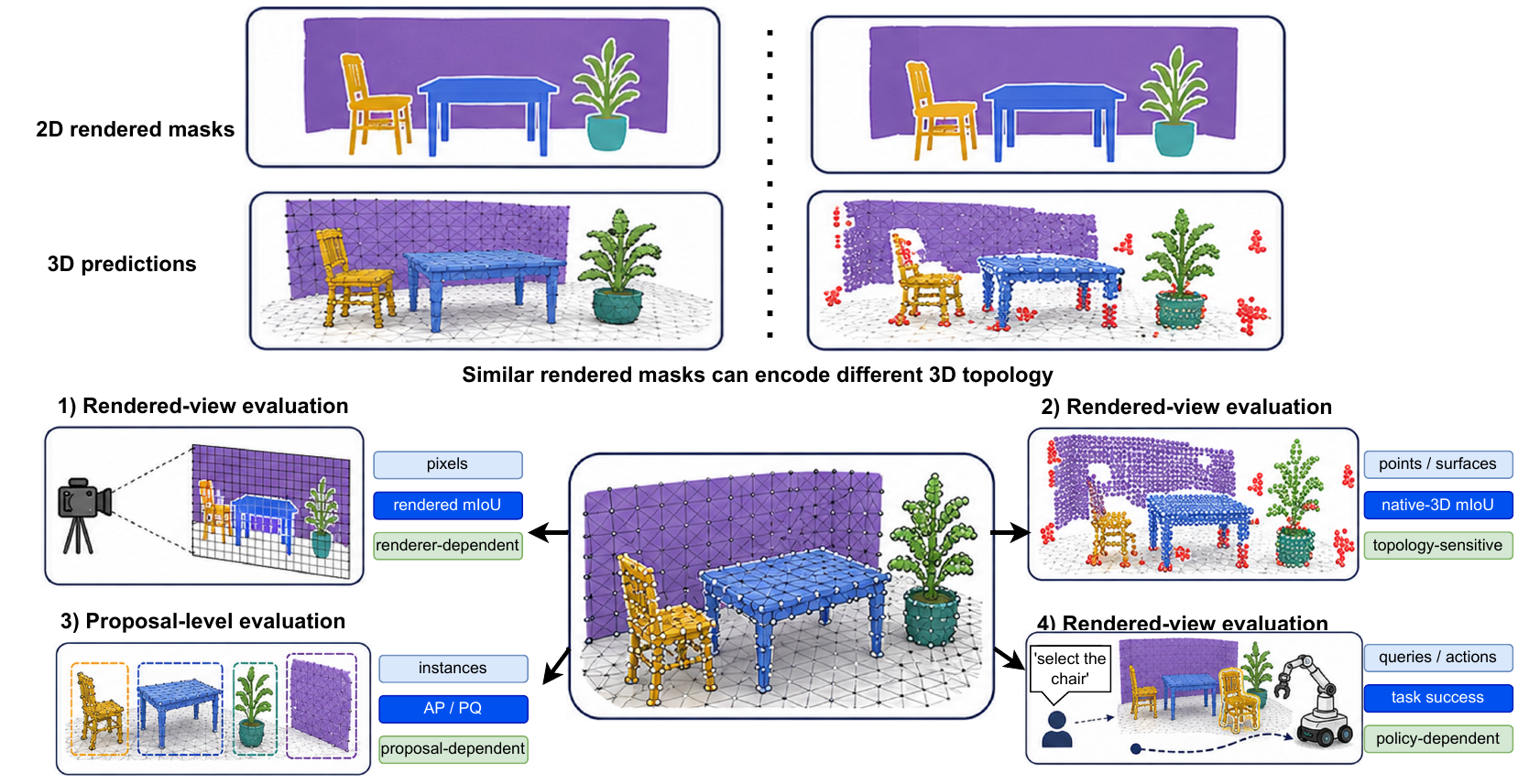}
\caption{Why rendered-view, native-3D, proposal-level, and downstream evaluations are not interchangeable. Rendered pixels can hide disconnected or incomplete 3D support; native-3D metrics expose geometry and topology; proposal metrics depend on instance generation and matching; and agent-level success additionally depends on the query and action policy. Cross-method comparison is valid only when geometry, views, prompts, conversions, and oracle access are controlled.}
\label{fig:evaluation-protocols}
\end{figure}

Figure~\ref{fig:evaluation-protocols} makes the output-contract distinction concrete. Two carriers can render nearly identical visible masks while differing in hidden-surface coverage, connectivity, proposal topology, or usefulness to an embodied policy. A metric becomes interpretable only after the evaluated object and its geometry, views, prompts, proposal source, and oracle access are fixed.

\subsection{The evaluation object comes before the metric}

Rendered-view evaluation compares a predicted image mask with a 2D annotation. It is natural for language fields such as LERF, 3D-OVS, LangSplat, and LEGaussians because their principal interface is differentiable rendering \citep{kerr_2023_lerf,liu_2023_3d_ovs,qin_2024_langsplat,shi_2024_legaussians}. InfoGS uses the same LERF-Mask images but exposes segmentation through decoder perturbation, and HumanCrafter reports novel-view accuracy for a fixed human-part vocabulary \citep{zhang_2025_infogs,pan_2025_humancrafter}. These are separate protocol families even though all report pixel IoU. The score depends jointly on semantic features, geometry, the renderer, and the chosen view. A carrier can therefore receive a high rendered score while remaining topologically incomplete on occluded surfaces.

Native-3D evaluation instead compares labels on points, mesh elements, voxels, or Gaussians. OpenScene and OpenNeRF are evaluated on Replica points in a matched OpenSeg setting, whereas OpenGaussian and COS3D include direct Gaussian-level protocols \citep{peng_2023_openscene,engelmann_2024_opennerf,wu_2024_opengaussian,zhu_2025_cos3d}. Even here, OpenScene benefits from cross-scene point-model pretraining while OpenNeRF optimizes a test-scene field. Converting a rendered method to native 3D requires sampling and thresholding; converting a native carrier to images introduces visibility and rasterization. These conversions are additional algorithms, not neutral display operations. Results should remain in separate rows unless the same conversion is applied to every method.

DesktopObjects-360 illustrates how one dataset can expose both domains. PointGauss introduced globally consistent rendered masks and native 3D instances for multi-object tabletop scenes, but its reported point-decoder results use a trained PTV3 model. CDSeg reuses the benchmark with propagated SAM2 masks and a renderer-vote carrier, while SAGOnline uses it for an online latency study rather than its principal NVOS/SPIn-NeRF accuracy protocol \citep{sun_2025_pointgauss,sun_2026_cdseg,sun_2025_sagonline}. Dataset name alone therefore does not align supervision, output, or timing boundaries.

Table 4 summarizes the minimum denominators for six evaluation families. Its purpose is not to prohibit cross-family analysis, but to require that any conversion, proposal oracle, prompt access, or downstream policy remain visible when results are compared.

\EvaluationProtocolTable

Instance segmentation adds a proposal source to the protocol. OpenMask3D classifies masks from a trained 3D proposal network; Open3DIS reports 2D-guided, 3D-network, and combined proposals; Details Matter varies image-only, point-only, and combined settings \citep{takmaz_2023_openmask3d,nguyen_2024_open3dis,jung_2025_details_matter}. OV3D-CG also reports oracle-mask experiments that isolate semantic reasoning from proposal quality \citep{zhou_2025_ov3d_cg}. These experiments answer valuable causal questions, but an oracle-mask AP must not be compared with end-to-end AP. The table row must specify proposal origin, whether ground-truth masks are used, and whether one or several class predictions are allowed per instance.

Newer results reinforce the need for protocol families. Rethinking OVRF and DiSCO-3D report rendered neural-field metrics; Any3DIS reports proposal AP; DITR reports closed-set point mIoU; ZeroPS reports object-part IoU; and SAM4D reports separate camera and LiDAR prompt segmentation \citep{lee_2025_rethinking_ovrf,petit_2025_disco3d,nguyen_2025_any3dis,abouzeid_2026_ditr,xue_2025_zerops,xu_2025_sam4d}. Placing these numbers in one ranking would erase the prediction object, vocabulary, supervision and sensor regime.

\subsection{Vocabulary and supervision define different generalization claims}

Base/novel protocols train with labels for a subset of classes and evaluate separately on seen and unseen categories. PLA, RegionPLC, and PGOV3D report harmonic IoU alongside base and novel mIoU over several partitions \citep{ding_2023_pla,yang_2024_regionplc,zhang_2025_pgov3d}. The partition must accompany every number: changing B15/N4 to B10/N9 changes both the amount of supervision and the semantic difficulty. Foreground-only metrics, which remove wall, floor, or ceiling, are also not closed-set mIoU under another name.

Annotation-free denotes the absence of manual target labels, not the absence of training. RegionPLC uses generated regional captions and multi-GPU point-language training; PGOV3D uses MLLM/SAM pseudo-entities, partial-view pretraining, and full-scene fine-tuning \citep{yang_2024_regionplc,zhang_2025_pgov3d}. MPEC curates entity masks and descriptions across several scene datasets \citep{wang_2025_mpec}. Their results measure reusable learned representations, whereas per-scene language fields optimize a carrier for the test scene. Both are valid paradigms, but they amortize cost and prior knowledge differently.

Free-form instructions define another task family. SOLE trains with visual, caption, and entity associations and evaluates category names as well as attribute or affordance descriptions \citep{lee_2025_sole}. OV3D-CG uses an MLLM and contextual views to classify proposals \citep{zhou_2025_ov3d_cg}. A CLIP label lookup, an attribute phrase, and a multi-step referring expression probe increasingly strong language capabilities. They require dedicated prompt sets and correctness criteria; success on a fixed category list does not imply compositional grounding.

\subsection{Cost must be reported stage by stage}

A complete ledger contains at least six entries: geometric reconstruction, 2D foundation-model preprocessing, association and reconciliation, carrier construction or training, persistent storage, and per-query inference. Omitting any stage changes the apparent conclusion. FlashSplat's closed-form segmentation avoids segmentation-field training, but still assumes a reconstruction and mask association \citep{shen_2024_flashsplat}. LEGaussians reports fast semantic rendering after dense feature extraction and per-scene optimization \citep{shi_2024_legaussians}. RegionPLC provides fast direct inference after regional-caption generation and dataset-scale training \citep{yang_2024_regionplc}. These are different allocations of work, not contradictory speed claims.

Table 5 operationalizes this ledger. Each row specifies what must be counted, whether the cost is paid per scene, observation, query, or amortized training run, and which omission most often makes a method appear cheaper than its actual deployment pipeline.

\CostLedgerTable

Query workload is equally important. Open-YOLO 3D avoids SAM and dense CLIP extraction by distributing detector labels across visible projections, reporting an efficient full-scene pass for a supplied vocabulary \citep{boudjoghra_2025_open_yolo3d}. For repeated new prompts on a fixed scene, a cached CLIP carrier can answer without rerunning the 2D detector and may therefore have lower marginal latency. Conversely, a dense language field may be wasteful for one known category list. Reports should distinguish first-query latency, subsequent-query latency, and the number of prompts or classes processed together.

Cross-scene and feed-forward systems require the training ledger as well. GOV-NeSF amortizes a neural semantic field across scenes, Ov3R jointly predicts geometry and semantics from video, and Uni3R removes supplied poses through a learned reconstruction representation \citep{wang_2024_gov_nesf,gong_2026_ov3r,sun_2026_uni3r}. HumanCrafter's roughly 6.2-second two-stage inference must be read beside seven days of training on eight A800 GPUs, specialized scans, and semantic annotation \citep{pan_2025_humancrafter}. Conversely, InfoGS reports a 16.3-minute shaping stage but presupposes a reconstructed 3DGS and multiple 2D foundation-model calls \citep{zhang_2025_infogs}. Deployment cost is meaningful only beside pretraining data, hardware, geometry source, and excluded preprocessing.

The same split is essential for latency. WildSeg3D's 5--20 ms interaction follows roughly 30 seconds of alignment and cache construction; Rethinking OVRF reports about 40 minutes of NeRF fitting plus 5 minutes of 3DGS transfer before fast rendering; Any3DIS reports hundreds of seconds for proposal generation \citep{guo_2025_wildseg3d,lee_2025_rethinking_ovrf,nguyen_2025_any3dis}. DITR and SAM4D move further cost into multi-GPU training \citep{abouzeid_2026_ditr,xu_2025_sam4d}. None of these numbers is contradictory once the measured stage is named.

Direct asset quality adds another evaluation domain. Lifting by Gaussians evaluates rendered views of extracted Gaussian assets, while Trace3D separately reports direct object-extraction quality and novel-view masks \citep{chacko_2025_lbg,shen_2025_trace3d}. RelationField adds scene-graph recall and relationship-guided instance IoU \citep{koch_2025_relationfield}. These protocols expose information ordinary mask mIoU hides, but cannot share one leaderboard.

Semantics-guided reconstruction further complicates baselines. COB-GS, ObjectGS, Trace3D and BEA-GS change the Gaussian geometry or its object-generating anchors, so they should not be compared with frozen-carrier methods without reporting the initial reconstruction and appearance change \citep{zhang_2025_cobgs,zhu_2025_objectgs,shen_2025_trace3d,mazzucchelli_2026_beags}. FMGS raw and SAM-refined results likewise represent different pipelines \citep{zuo_2025_fmgs}.

Large-scene and articulated systems expose different missing denominators. DiLEGS reports mask preprocessing, semantic training, carrier size and storage, while FreeArtGS decomposes part segmentation, joint estimation and end-to-end optimization \citep{li_2026_dilegs,dai_2026_freeartgs}. NG-GS reports hardware and boundary gains but not a complete end-to-end runtime \citep{he_2026_nggs}; comparisons should mark that cell as unknown rather than infer efficiency from representation choice.

LUDVIG illustrates why `\texttt{learning-free'' is not }`cost-free'': its uplifting is brief relative to learned fields, yet the same experiment still requires pretrained 3DGS geometry and substantial 2D feature extraction; graph diffusion also moves work to inference \citep{marrie_2025_ludvig}. Proto-SaGa reports 30K iterations and hardware but no wall-clock, SAM preprocessing or carrier size \citep{oh_2026_proto_saga}. A complete table must preserve those unknowns rather than equate iteration count with elapsed cost.

The same accounting changes how recent discrete Gaussian interfaces should be described. SAGOnline reports immediate first-mask interaction, about 1.47 seconds to aggregate 207 propagated masks, and 27 ms refined rendering, but assumes a reconstructed 3DGS and a video foundation-model pass. CDSeg processes one to seven million point-completed Gaussians in roughly 2.23--3.74 seconds for 300 views, explicitly excluding external mask inference and prior reconstruction \citep{sun_2025_sagonline,sun_2026_cdseg}. These are useful stage-level measurements, not reconstruction-inclusive end-to-end runtimes.

Feed-forward systems make the amortization boundary especially visible. PanSt3R reports about 2.3 minutes for direct multi-view panoptic prediction but about 35 minutes when LUDVIG supplies novel-view 3DGS output \citep{zust_2025_panst3r}. PartField reports a sub-second field pass and roughly ten seconds including clustering only after two weeks of training on eight A100 GPUs \citep{liu_2025_partfield}. CADRNet reports 372.7 GFLOPs per query, of which 332.1 belong to frozen CLIP/Vicuna components \citep{yu_2026_cadrnet}. These values should not be collapsed into one speed ranking: they measure different outputs and allocate construction cost at different scopes.

\subsection{Minimum reporting checklist}

Every quantitative row should state: dataset and version; category split and vocabulary access; output domain; geometry source; proposal source; prompt protocol; supervision type; and whether the reported runtime includes reconstruction, 2D inference, fusion, and query. Memory should separate transient GPU usage from persistent scene storage. ``Real-time,'' ``training-free,'' ``annotation-free,'' and ``generalizable'' should be treated as claims requiring this ledger, not as self-explanatory method labels.

\subsubsection{Part, prompt, and agent workloads}

Part methods expose prompt and amortization mismatches. Find3D and PatchAlign3D report sub-second point inference only after large SAM/VLM/DINO pseudo-label pipelines and cross-shape training \citep{ma_2025_find3d,hadgi_2026_patchalign3d}. GeoSAM2 reports about 30 seconds but receives a precise visible 2D prompt and uses eight A800 GPUs for adaptation \citep{deng_2026_geosam2}. MV3DIS reports strong scene AP without a complete mask-generation or wall-clock ledger \citep{zhao_2026_mv3dis}. These entries require separate rows rather than a single part-segmentation speed column.

The same family contains three incompatible meanings of efficiency. PartSLIP explicitly incurs long runtime from rendering and repeated GLIP calls but gives no normalized wall-clock; COPS is training-free yet still renders 48 views and runs DINOv2, clustering and optional CLIP labeling; CoSMo3D reports 0.9 seconds per shape only after canonical-corpus construction and cross-shape training \citep{liu_2023_partslip,garosi_2025_cops,jin_2026_cosmo3d}. Training-free, direct inference and end-to-end cheap are separate claims.

Reasoning segmentation adds language-agent latency. REALM reports 8.68 seconds per query on an RTX 3090, decomposed into 2.53 seconds for global MLLM reasoning, 2.48 for local reasoning and 3.67 for 50 refinement steps \citep{shi_2026_realm}. Its reported 354.72 FPS measures rendering rather than the complete query. This is precisely why throughput numbers must name their boundary; otherwise a fast carrier renderer can obscure a much slower agentic interface.

Workshop evidence adds useful mechanisms but should remain visibly lower-weight than full conference evidence. SAP reports twelve 1024-pixel renders and one RTX 4060 but no end-to-end time; DiscoNeRF separates NeRF reconstruction from object-field optimization without reporting wall-clock or storage \citep{bai_2025_sap,dumery_2025_viewconsistent}. Their headline metrics also occupy different rows: primitive AP on native point clouds versus class-agnostic masks rendered from a field. Neither supports a direct rank against semantic part mIoU or scene-instance AP.

Scene-specialized ledgers remain incomplete in different ways. Fun3DU and modality-guided transfer omit complete hardware/wall-clock accounting; Wheat3DGS reports reconstruction but scores rendered masks before trait extraction; DCSEG omits stage times \citep{corsetti_2025_fun3du,spoecklberger_2025_modality,zhang_2025_wheat3dgs,wiedmann_2025_dcseg}. OpenSplat3D separates 20--45 minutes of instance fitting, clustering, and language aggregation, so only it supports a stage-level latency claim \citep{piekenbrinck_2025_opensplat3d}.

Evaluation domain must also remain attached to the number. OpenSplat3D reports rendered LERF-mask mIoU/mBIoU of 84.0/78.8 and native ScanNet++ class-agnostic AP/AP50/AP25 of 24.5/41.7/57.1; these are complementary tests, not two estimates of the same accuracy \citep{piekenbrinck_2025_opensplat3d}. Wheat3DGS's 0.50 rendered-mask IoU supports multi-view consistency but cannot establish native 3D topology \citep{zhang_2025_wheat3dgs}. DCSEG's selected-scene ScanNet comparison likewise depends on using only 200 images \citep{wiedmann_2025_dcseg}. Every result should preserve its observation budget and output domain.

RangeSAM adds a different foundation-transfer denominator. Its reported 60.9 SemanticKITTI validation mIoU is below RangeFormer's 73.3, but the experiment tests whether SAM2 Hiera initialization can survive a modality shift, not whether an open vocabulary is exposed \citep{kuhn_2026_rangesam}. The model uses eight 40-GB A100 GPUs and 60 epochs of nuScenes pretraining before supervised target training; without wall-clock and inference latency, the claim is architectural viability rather than efficiency.

\subsubsection{Online maps, hierarchies, and graphs}

MaskClustering and SAI3D report native point-instance AP, yet their training-free label excludes scene-wide 2D inference and graph construction \citep{yan_2024_maskclustering,yin_2024_sai3d}. SAI3D explicitly notes cost grows linearly with image count; MaskClustering omits normalized hardware and time. PS3 further adds Mask3D proposals, Semantic-SAM masks, SAM2.1-Hiera-Large tracking and SigLIP features while reporting no complete ledger \citep{yen_2026_ps3}. Their AP values should be accompanied by view sampling and every upstream foundation call.

Hierarchical systems expose the same boundary problem. OVIR-3D's 30-fps fusion/20-ms retrieval excludes reconstruction and Detic; N2F2 omits field-training time and storage; Ultrametric Feature Fields gives no normalized ledger \citep{lu_2023_ovir3d,bhalgat_2024_n2f2,he_2024_ultrametric}. Search3D separates object/part construction, SigLIP aggregation, vocabulary embedding, and final similarity search \citep{takmaz_2025_search3d}. A fast lookup is not evidence of a cheap persistent representation.

Three 2024 mapping systems reinforce both the metric and cost boundaries. ConceptGraphs reports 40.63 mAcc and 35.95 F-mIoU on Replica but gives no normalized construction time and acknowledges multiple LVLM/LLM calls. Open-Fusion reports 50 FPS for geometry and 4.5 FPS with semantics, together with 0.62 mAcc and 0.59 F-mIoU on 20 ScanNet scenes; the semantic teacher is the bottleneck. HOV-SG reports 1,493 MB over eight HM3DSem scenes versus 6,068 MB for its adapted VLMaps baseline, yet explicitly calls graph construction too slow for real-time mapping \citep{gu_2024_conceptgraphs,yamazaki_2024_openfusion,werby_2024_hovsg}. These numbers measure different datasets, carriers and stages: none supports a three-way efficiency ranking.

Online mapping, point-backbone training, and functional graphs are incompatible protocols. OVI-MAP scores frame-sampled online AP, CUA-O3D a 50-epoch point model, and four graph systems node or triplet retrieval under different annotations and LLMs \citep{deng_2026_ovimap,li_2025_cua_o3d,zhang_2025_openfungraph,werby_2026_keysg,kim_2026_op3dsg,fu_2026_funfact}. Output type, oracle access, frame sampling, and preprocessing must stay visible.

Scale changes the unit: LOST-3DSG estimates persistent bytes, KNA-SG reports query latency, and OpenCity3D reports carrier construction for 10k images \citep{ferraina_2026_lost3dsg,xu_2026_knasg,bieri_2025_opencity3d}. They cannot share one unlabeled efficiency column.

DovSG measures local update latency, HAECcity amortized feed-forward inference after an incompletely timed pseudo-label engine, and Open-World 3DSG retrieval without end-to-end time or index size \citep{yan_2025_dovsg,rusnak_2025_haeccity,yu_2026_ow3dsg}. These remain separate ledger rows.

OpenVoxel reports about three minutes for group and scene-map construction and below one second per query on one RTX 5090, but starts from a pretrained sparse voxel reconstruction and runs SAM2 plus captioning models \citep{huang_2026_openvoxel}. The result is a useful staged measurement, not a reconstruction-inclusive end-to-end runtime.

Ref-LERF also spans incompatible training regimes. ReferSplat reports 58 minutes of per-scene, expression-conditioned training, 26.8 FPS and 3.3 MB for one A6000 scene, whereas GaussDet evaluates referring expressions through zero-shot 2D detector votes after 30k instance-feature steps \citep{he_2025_refersplat,hassan_2026_gaussdet}. A table should label whether benchmark expressions generated training pseudo-masks, whether the detector was frozen, and whether grouping and reconstruction costs are counted. Otherwise a higher referring mIoU can be mistaken for stronger unseen-expression transfer.

ZeroSplat sharpens the distinction between training-free and low-latency \citep{ding_2026_zerosplat}. It reports no semantic optimization and zero extra semantic-feature storage, yet a complete generalized Teatime query takes about 2,306 seconds on an RTX 4090D because VLM parsing, SAM3 masks, lifting and refinement occur at query time. TrackRef3D instead amortizes automatic detection, tracking and caption generation into a five-epoch scene field, but reports neither wall time nor storage \citep{tan_2026_trackref3d}. Both need stage-complete reporting before efficiency conclusions.

\subsubsection{Compact and dynamic carriers}

Compact carriers provide unusually concrete but still partial ledgers. LightSplat reports roughly four seconds for semantic distillation, 0.002 seconds per ScanNet text query and two bytes per Gaussian; SCOUP reports 0.5--1.4 minutes for coefficient uplift, 15 FPS and a 0.1-GB sparse carrier in one ablation \citep{bang_2026_lightsplat,budimir_2026_scoup}. ProFuse reports about 2.5 minutes of initialization, 15 minutes of geometry and 0.5 minutes of semantics \citep{chiou_2026_profuse}. None of these figures includes an identical base reconstruction and 2D preprocessing boundary, so the carrier measurement is valuable without supporting a universal speed ranking.

Dynamic evaluation needs temporal denominators. 4D LangSplat, LangField4D, ST4R-Splat and 4D Synchronized Fields report combinations of frame mIoU, temporal accuracy, video IoU and interval IoU on extensions of HyperNeRF or Neu3D \citep{li_2025_4dlangsplat,xu_2026_langfield4d,meng_2026_st4rsplat,barhdadi_2026_4dsync}. Each result must state which expressions and intervals were human annotated, which captions trained the field, whether the target identity was supplied, and whether the method is peer reviewed. TIBR4D instead measures object extraction after a chosen number of iterations, with dynamic reconstruction explicitly outside the timer \citep{wu_2026_tibr4d}.

Object-mask 4D benchmarks are also not a single denominator. SA4D, TRASE, SGG, Split4D and Multi4D evaluate different annotated extensions, camera regimes and reconstruction backbones; some require tracked target IDs while others cluster independent masks \citep{ji_2024_sa4d,li_2026_trase,wei_2026_sgg,hu_2025_split4d,wang_2026_multi4d}. Their ledgers range from SA4D's roughly 30-minute field plus sub-10-second refinement, through TRASE's 32-minute semantic pipeline, to Multi4D's 1.2-hour joint reconstruction. DGD's roughly five-hour high-dimensional distillation and 4-LEGS's five-hour autoencoder plus per-timestep fitting measure still different carriers \citep{labe_2024_dgd,fiebelman_2025_4legs}. A speed claim must state whether reconstruction, masks, feature extraction, carrier learning and clustering are counted.

\subsubsection{Specialized endpoints remain separate protocols}

Several specialized records reinforce why protocol families must remain separate. 3DIML reports renderer-oriented PQscene and a scene-specific minute ledger; SAM2Object reports native point-instance AP after sequence tracking; EmbodiedSAM, ESAM++ and MoonSeg3R report online AP with different sensor and teacher assumptions \citep{tang_2024_3diml,zhao_2025_sam2object,xu_2025_embodiedsam,liu_2026_esampp,du_2026_moonseg3r}. GeoCGA scores referring masks after language expansion, TEXTRIX scores generated mesh parts, and 100Editor primarily scores the downstream edit \citep{tao_2026_geocga,zeng_2026_textrix,wu_2026_100editor}. A common metric name cannot erase these different prediction objects.

\section{Failure Propagation and Framework Diagnostics}

Most errors in 2D-to-3D foundation-model transfer are compositional: an upstream mistake changes the evidence available to every downstream stage. A correct text embedding cannot recover an object that was never proposed, and a perfect association rule cannot repair a geometrically incorrect reconstruction. We organize failures by the first stage at which information is lost, then trace how the error propagates through the persistent carrier.

IQGS and OV3DSeg-VGGT expose complementary structural limits. IQGS can exhaust its fixed query bank or overfit query indices to scene-specific IDs. OV3DSeg-VGGT can learn consistent semantics, yet lift them through erroneous small-object depth and oversmooth boundaries with neighborhood consensus \citep{gao_2026_iqgs,zhou_2026_ov3dseg_vggt}. Query-capacity sweeps and depth-oracle/KNN ablations isolate these failures more cleanly than aggregate mIoU alone.

\begin{figure}[H]
\centering
\includegraphics[width=\textwidth]{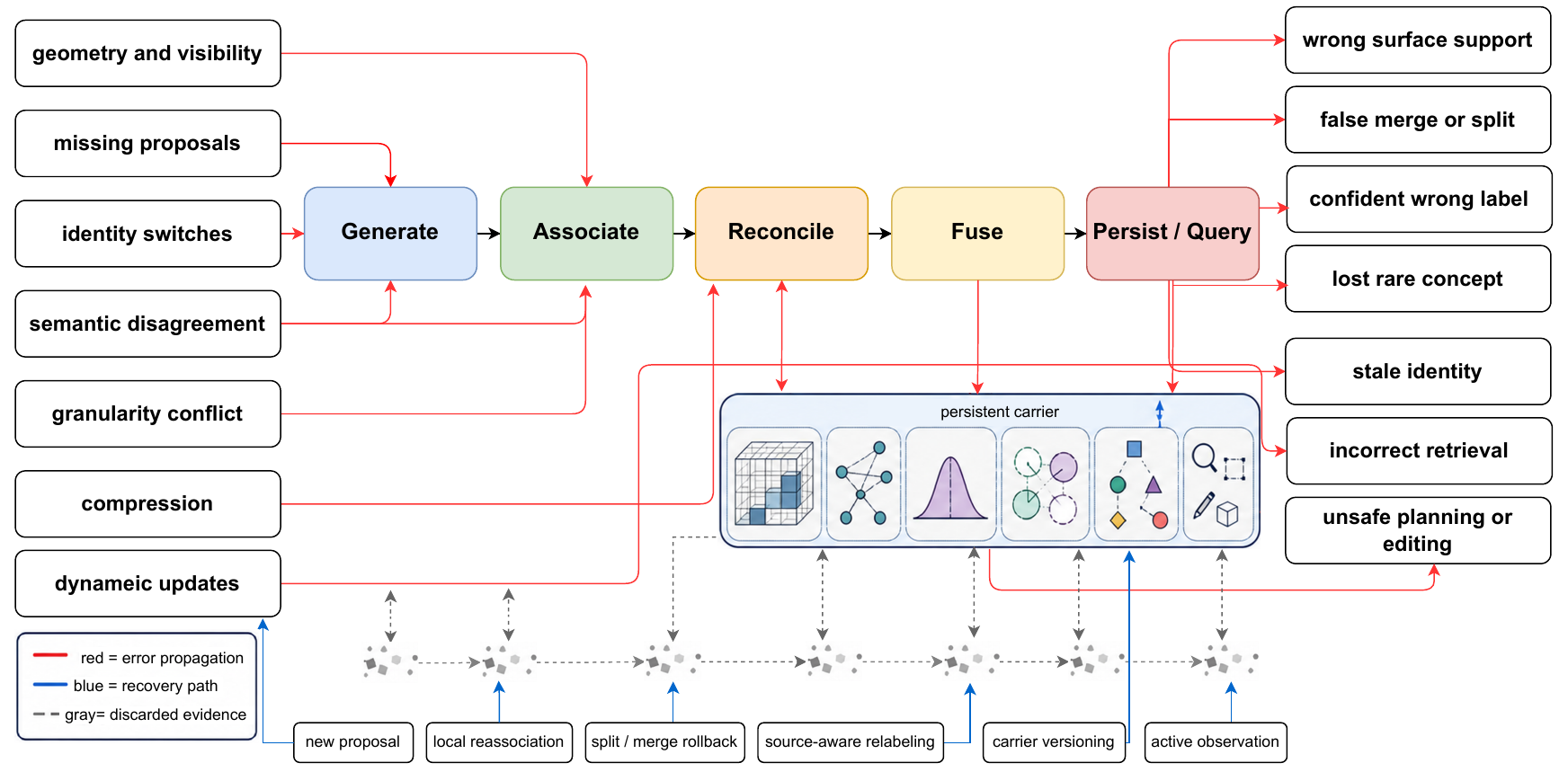}
\caption{Failure propagation, irreversible commitment, and recovery operations in LAF. Red paths show how upstream errors can alter persistent state and downstream behavior; gray paths denote discarded evidence; and blue paths denote recovery operations supported when sufficient uncertainty and provenance remain. The first irreversible loss is system- and query-dependent and may occur at any operator where evidence needed for a later correction is discarded.}
\label{fig:failure-recovery}
\end{figure}

Figure~\ref{fig:failure-recovery} visualizes how geometry, proposals, identity, semantics, granularity, compression, and updates can enter at different operators and become persistent downstream errors. The diagram deliberately does not assign a universal first-loss stage: depending on the system and query, an irreversible commitment may occur during Generate, Associate, Reconcile, Fuse, or Persist/Query.

Table 6 provides a compact diagnostic map from each failure source to its propagation path, a causal test, and the recovery operation that the carrier would need to support. The organizing rule is to diagnose the first irreversible loss rather than attributing every downstream error to the final classifier.

\FailureDiagnosticTable

\subsection{Geometry and visibility errors corrupt support}

Calibrated projection assumes accurate poses, depth, and synchronization. A depth tolerance that is too strict drops valid observations; one that is too loose transfers a foreground mask to the surface behind it. Point-based systems such as OpenScene and Open3DIS make this policy explicit through visibility tests and superpoint support \citep{peng_2023_openscene,nguyen_2024_open3dis}. Renderer-based systems soften the assignment, but do not remove the error. In NeRF and 3DGS carriers, floaters, blurred density, or splats crossing physical boundaries spread semantic gradients to the wrong support \citep{kerr_2023_lerf,ye_2024_gaussian_grouping}. OpenNeRF can request rendered views for high-variance regions, but an uncertainty rule conditioned on the same imperfect geometry may repeatedly observe an occluded surface or reinforce a rendering artifact \citep{engelmann_2024_opennerf}.

Learned reconstruction couples the failure modes more tightly. Ov3R and Uni3R can remove known poses and sensor depth, but an incorrect predicted point map changes both geometry and the location of its semantic descriptors \citep{gong_2026_ov3r,sun_2026_uni3r}. PointGS introduces a temporary Gaussian carrier and later registers its clusters back to the original point cloud; a registration error can transfer a coherent cluster to the wrong points \citep{song_2026_pointgs}. The practical lesson is to retain geometric uncertainty or alignment confidence rather than presenting the final semantic score as independent evidence.

\subsection{Proposal topology imposes a hard ceiling}

Open-vocabulary classifiers name surviving proposals; they generally do not split a merged mask or recreate an omitted small object. OpenMask3D exposes this ceiling directly because classification follows a fixed 3D proposal network \citep{takmaz_2023_openmask3d}. Open3DIS increases coverage by combining 2D-guided and learned 3D proposals \citep{nguyen_2024_open3dis}. Details Matter further removes 2D overlaps, tracks co-visible superpoints, prunes inconsistent support, merges partial instances, and removes contained duplicates \citep{jung_2025_details_matter}. Each repair improves one topology failure while risking another: overlap removal fragments instances, iterative merging can join neighbors, and inclusion removal can erase a legitimate small object.

Fast labeling systems inherit the same ceiling. Open-YOLO 3D distributes detector labels over Mask3D proposals and avoids expensive dense features, but a small object missing from the 3D proposals cannot be recovered by multi-view voting \citep{boudjoghra_2025_open_yolo3d}. SOLE lets semantics influence proposal generation itself \citep{lee_2025_sole}. This relaxes the class-agnostic ceiling but creates a new bias: concepts emphasized by captions and entity supervision can acquire better mask topology than unmentioned or ambiguous objects.

\subsection{Identity failures become persistent state transitions}

External trackers can create stable but wrong identities. Gaussian Grouping and Segment then Splat efficiently persist tracker-defined objects, yet an identity switch or early merge becomes a property of the 3D carrier \citep{ye_2024_gaussian_grouping,lu_2025_segment_then_splat}. In InfoGS the same error shapes which Gaussians respond together, so later perturbations can consistently select or move multiple objects as one \citep{zhang_2025_infogs}. Geometry-based memory systems such as SAM3D, Gaga, and OnlineAnySeg delay identity until observations overlap in 3D, but append-only or thresholded merges can still be irreversible \citep{yang_2023_sam3d,lyu_2026_gaga,tang_2025_onlineanyseg}.

Late semantic attachment does not remove teacher error; it localizes it. DCSEG can recover a coherent tail-class cluster that OVSeg or OpenSeg cannot name, while OpenSplat3D can retrieve a whole instance whose boundary splats leaked through alpha blending or specular reflection \citep{wiedmann_2025_dcseg,piekenbrinck_2025_opensplat3d}. Wheat3DGS exposes the analogous domain-specific ceiling: a reliable but lower-recall detector produces cleaner supervision, yet missing heads bias later counting and phenotyping \citep{zhang_2025_wheat3dgs}. Proposal quality, semantic correctness and geometric membership should therefore be audited separately.

Consensus can turn correlated teacher bias into confidence. MaskClustering assumes most views segment an object correctly, so repeated under- or over-segmentation can distort supporter counts despite its filtering \citep{yan_2024_maskclustering}. SAI3D's progressive thresholds reduce early merge chains but cannot repair a pose or SAM error that created the affinity \citep{yin_2024_sai3d}. PS3 replaces majority masks with a tracked masklet, trading independent-view inconsistency for pivot-mask granularity and drift \citep{yen_2026_ps3}. Robust reconciliation should retain contradictions rather than only the winning edge or track.

Projection-based backbone transfer has a different loss channel. RangeSAM keeps only the nearest LiDAR return per range pixel and later assigns dense point labels through k-nearest-neighbor voting \citep{kuhn_2026_rangesam}. A reported point mIoU therefore mixes encoder performance with raster collision and postprocessing. These errors should not be grouped with camera-pose failures simply because both pipelines contain a projection operator.

Graph carriers increase the blast radius. Octree-Graph stores occupancy, semantics, and relations on object nodes, while ReLaGS and OGScene3D support relational or confidence-aware scene queries \citep{wang_2025_octree_graph,xie_2026_relags,zhu_2026_ogscene3d}. If two objects are merged, the error changes not only a mask but also captions, edges, retrieval answers, and navigation occupancy. Such systems need merge provenance, edge confidence, and local graph repair rather than only a final object ID.

\subsection{Semantic disagreement can be averaged into confidence}

Multi-view mean pooling hides whether observations agree. Viewpoint, occlusion, and the 2D teacher's category bias can produce incompatible embeddings that collapse to a plausible average. LEGaussians learns uncertainty-aware smoothing, VALA weights visibility, and Octree-Graph favors views that distinguish an instance from its neighbors \citep{shi_2024_legaussians,wang_2025_vala,wang_2025_octree_graph}. These mechanisms are preferable to unqualified averaging, but their confidence is still conditional on the available observations.

Generated text creates a related supervision failure. PLA, RegionPLC, MPEC, and PGOV3D use captions, entities, or MLLM-generated vocabularies to train point representations \citep{ding_2023_pla,yang_2024_regionplc,wang_2025_mpec,zhang_2025_pgov3d}. A hallucinated entity, omitted object, or ambiguous regional caption becomes a training pair. Multi-view agreement helps only when the error varies across views; systematic teacher bias can be reinforced. Provenance, source-specific confidence, negative evidence, and pseudo-label revision should be first-class training data.

\subsection{Granularity and language create valid but incompatible answers}

A part, object, and object group may all satisfy the same image mask or phrase at different scales. LangSplat fixes three levels, while LERF, GARField, SAGA, and S2AM3D expose scale as a continuous or prompted variable \citep{qin_2024_langsplat,kerr_2023_lerf,kim_2024_garfield,cen_2025_saga,su_2026_s2am3d}. Continuous control avoids one fixed hierarchy but does not guarantee semantically valid transitions. Missing 2D groups and spatially disconnected semantic similarity still lead to unstable masks. HumanCrafter exposes a different coverage ceiling: frozen DINOv2 supplies rich features, but the supervised head predicts 28 predefined human parts \citep{pan_2025_humancrafter}. Foundation transfer and open-vocabulary access are therefore separate claims.

Contextual language reasoning adds another ambiguity. OV3D-CG can use room context to disambiguate an instance, while COS3D sends language relevance back into a boundary-aware instance field \citep{zhou_2025_ov3d_cg,zhu_2025_cos3d}. Both can amplify a plausible but incorrect initial interpretation. Robust systems should support abstention, show which geometry or views justify the answer, and test counterfactuals in which context is removed or changed.

These carriers add recognizable cascades. NeRF-to-3DGS transfer can alter semantic support; single-view SceneDINO can hallucinate occluded completion; Any3DIS and PointSeg can lock an early prompt or tracking error into proposal topology; WildSeg3D can reuse a bad alignment through its cache; and SAM4D can propagate a temporal error across camera and LiDAR \citep{lee_2025_rethinking_ovrf,jevtic_2025_scenedino,nguyen_2025_any3dis,he_2025_pointseg,guo_2025_wildseg3d,xu_2025_sam4d}. These are not isolated corner cases: they follow directly from what each system chooses to persist.

Geometry correction can itself overfit noisy semantic evidence. A false boundary may cause COB-GS to over-split, a wrong voted ID may steer ObjectGS anchor growth, and a hidden-occupancy error may move a BEA-GS splat with little image evidence \citep{zhang_2025_cobgs,zhu_2025_objectgs,mazzucchelli_2026_beags}. Such methods need rollback and appearance-preservation tests, not only better final mask scores.

\subsection{Hierarchies and graphs amplify upstream errors}

Hierarchy adds correlated rather than independent errors. N2F2 can choose the wrong semantic scale across many pixels, an ultrametric bottleneck can move an entire subtree, Search3D can inherit one normal-based fracture across every part query, and OVIR-3D can preserve an early merge decision throughout online mapping \citep{bhalgat_2024_n2f2,he_2024_ultrametric,takmaz_2025_search3d,lu_2023_ovir3d}. A hierarchy-aware diagnostic should therefore perturb scale, threshold, parent assignment and observation order, then report which outputs remain stable rather than evaluating only one selected partition.

Object graphs and online dictionaries widen this test. ConceptGraphs can reuse one missed, duplicated or wrongly captioned node across relations and plans; Open-Fusion can average a mistaken region-channel assignment into later TSDF observations; HOV-SG can propagate an incorrect floor or room parent through every constrained query \citep{gu_2024_conceptgraphs,yamazaki_2024_openfusion,werby_2024_hovsg}. Evaluation should therefore replay observations in different orders, perturb association thresholds, and require local rollback without rebuilding the entire map.

Functional graphs add an ontology cascade. OpenFunGraph and FunGraph cannot create an action anchor that the 2D part proposal omitted; OP3DSG can recover candidates through curated object-part knowledge but inherits that ontology's blind spots; FunFact can calibrate uncertain edges but not repair a missing or wrongly parented node \citep{zhang_2025_openfungraph,rotondi_2025_fungraph,kim_2026_op3dsg,fu_2026_funfact}. OVI-MAP and KeySG expose the complementary evidence-selection failure: sparse semantic views or keyframes may preserve geometry yet omit the observation needed by a later phrase \citep{deng_2026_ovimap,werby_2026_keysg}.

\subsection{Dynamic, compressed, and generative carriers require recovery tests}

Compression and scale introduce two further cascades. LOST-3DSG can alias visually similar moving objects after discarding dense evidence; KNA-SG can poison later queries by persisting one mistaken verified relation; OpenCity3D can turn imagery age, view angle or demographic imbalance into coherent city-wide heatmaps \citep{ferraina_2026_lost3dsg,xu_2026_knasg,bieri_2025_opencity3d}. Lifecycle provenance, relation rollback and dataset/fairness audits are therefore carrier requirements, not optional downstream safeguards.

Persistent update layers create their own cascade. DovSG can leave stale CLIP evidence after motion; HAECcity can distill a synthetic-view coverage error into a reusable network; Open-World 3DSG can return a stale vector chunk after its source node or edge changes \citep{yan_2025_dovsg,rusnak_2025_haeccity,yu_2026_ow3dsg}. Dynamic-map tests should therefore exercise invalidation, pseudo-label revision and transactional reindexing rather than only fresh construction.

OpenVoxel makes another irreversible boundary explicit: sequential mask merging fixes a group before canonical captioning compresses its appearance into text \citep{huang_2026_openvoxel}. Evaluation should reorder views and test paraphrases or counterfactual descriptions; otherwise grouping order and caption coverage remain hidden behind final query mIoU.

Expression-aware and detector-based Gaussian methods fail at different interfaces. ReferSplat can leak benchmark expressions into scene-specific pseudo-supervision, so it needs frozen-carrier tests on unseen expressions and scenes \citep{he_2025_refersplat}. GaussDet can aggregate many consistent semantic votes over an incorrectly merged group, producing high confidence without correct topology \citep{hassan_2026_gaussdet}. Detector-oracle, grouping-oracle and end-to-end ablations are needed to identify which boundary actually limits the result.

Referring pipelines make semantic derailment observable earlier. ZeroSplat fails when dense clutter corrupts a VLM box or when the parsed category misstates the instruction; all later masks and lifted labels then follow the wrong target \citep{ding_2026_zerosplat}. TrackRef3D can instead poison a field through an identity switch or an over-broad synonym cluster \citep{tan_2026_trackref3d}. Both should expose intermediate boxes, tracks and canonical labels, and evaluate correction without rebuilding the entire scene.

Dynamic carriers add three failure axes. A tracked identity can switch under deformation, a continuous action can cross a poorly localized temporal boundary, and a motion cue can correlate with the wrong semantic state \citep{li_2025_4dlangsplat,xu_2026_langfield4d,barhdadi_2026_4dsync}. ST4R-Splat reduces spatial--temporal entanglement by grounding the instance before state lookup, but still depends on its tracks and generated descriptions \citep{meng_2026_st4rsplat}. TIBR4D can refine boundaries to convergence yet cannot split or merge the initial identity topology \citep{wu_2026_tibr4d}. Evaluation should therefore perturb tracks, action boundaries and topology separately.

Carrier choice creates additional cascades. A tracker switch contaminates SA4D's identity table or SGG's explicit trajectory; a low-resolution teacher map makes DGD's large semantic carrier confidently coarse; an absent action can still activate 4-LEGS; and a transient primitive included as semantic support can fragment identity \citep{ji_2024_sa4d,wei_2026_sgg,labe_2024_dgd,fiebelman_2025_4legs}. TRASE and Split4D avoid global tracked IDs but can merge small overlapping objects during deferred feature reconciliation \citep{li_2026_trase,hu_2025_split4d}. Multi4D addresses the transient-primitive case structurally, yet still inherits SAM-mask and clustering errors \citep{wang_2026_multi4d}. Stress tests should separately alter mask identity, teacher resolution, absent queries, clustering density and persistent/transient allocation.

Compression failures deserve query-after-compression tests. LightSplat can discard a weakly supported mask before assigning its cluster ID, SCOUP can remove a rare code atom with Top-K filtering, and ProFuse can pool mismatched masks into one proposal \citep{bang_2026_lightsplat,budimir_2026_scoup,chiou_2026_profuse}. Aggregate mIoU may hide these losses when common objects dominate. Rare-query recall, code occupancy, reversible cluster splitting and held-out paraphrases should be reported alongside storage savings. Online methods add state-transition failures. A SAM2 track can propagate one wrong object through a superpoint graph; a query matcher can irreversibly merge two identities; an opaque reconstructive state can make a wrong association appear temporally stable \citep{zhao_2025_sam2object,xu_2025_embodiedsam,du_2026_moonseg3r}. Compression in ESAM++ changes latency and geometry capacity but not those logical failure modes \citep{liu_2026_esampp}. Recovery tests should inject track switches, temporary disappearance and contradictory later evidence.

Relational and generative carriers fail earlier in their ontology. GeoCGA cannot align a relation to an object missing from the proposal graph, while TEXTRIX may reproduce a coherent partition created by its DINO-based pseudo-target pipeline rather than a human part taxonomy \citep{tao_2026_geocga,zeng_2026_textrix}. These are not ordinary boundary errors: the available entity or part inventory is wrong before the final mask is rendered.

\subsection{Practical design rules}

The recurring lessons are simple but demanding: preserve uncertainty until the query requires a decision; store merge and supervision provenance; separate object discovery from naming in evaluation even when the model couples them; provide a small-object recovery path; and test the carrier under repeated, relational, and corrective queries. Most importantly, diagnose the first irreversible loss of information. Downstream sophistication should not be credited with repairing evidence that the pipeline no longer retains.

\section{From Passive Segmentation to Agentic 3D Perception}

Most benchmarks ask a carrier to answer one segmentation query after observing a static scene. An embodied agent faces a different problem: it must retain identities over time, choose what to observe next, revise beliefs after change, and use segmented entities in relational tasks. The methods reviewed here do not yet constitute a complete agentic perception stack, but several mechanisms mark the transition from passive scene labeling to persistent spatial memory.

\begin{figure}[H]
\centering
\includegraphics[width=\textwidth]{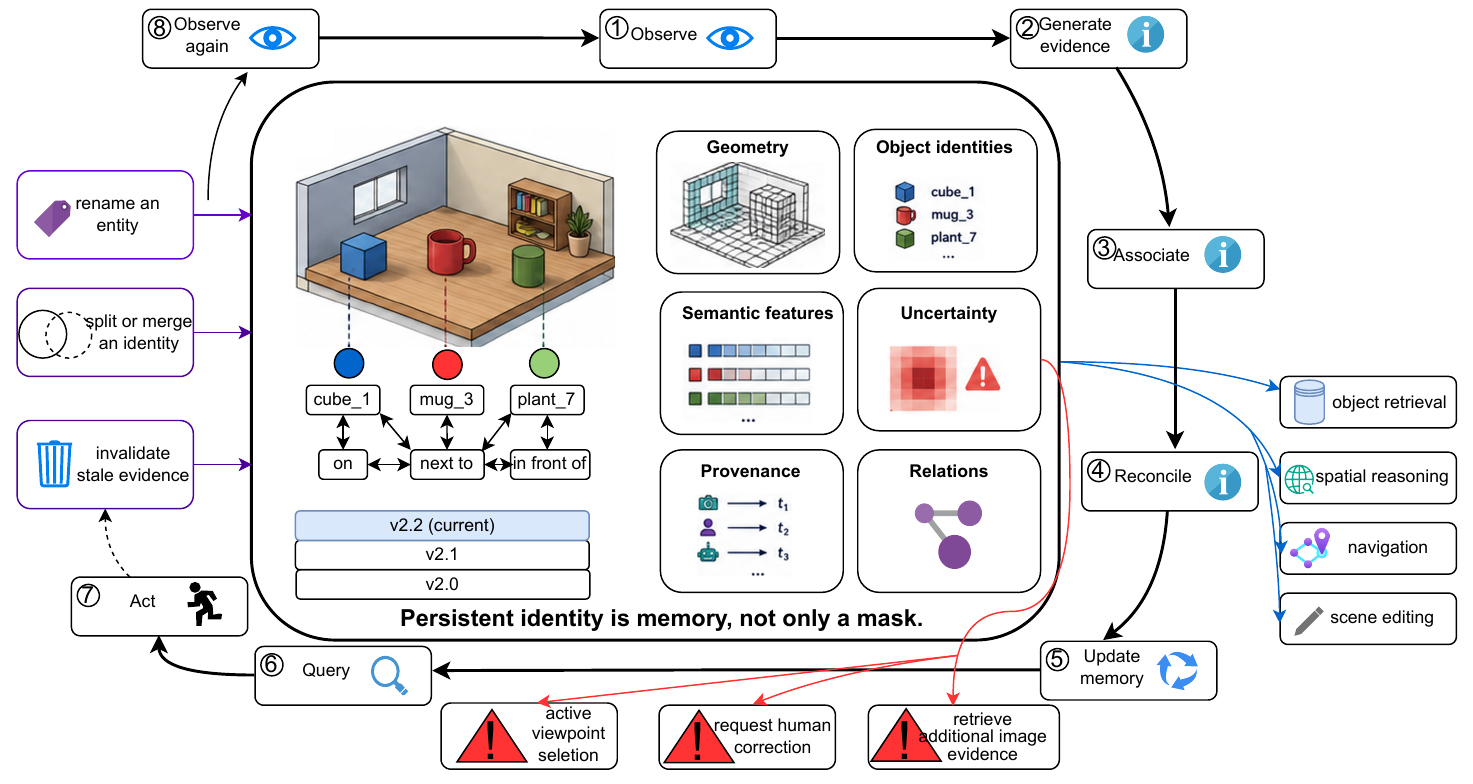}
\caption{From passive segmentation to agentic 3D perception. A versioned carrier retains geometry, identities, semantic state, uncertainty, provenance, and relations while observations, queries, actions, and corrections form a closed loop. In this schematic, Update memory combines the Fuse and Persist operations. Uncertainty may trigger an active viewpoint, human correction, or retrieval of additional image evidence, while explicit revision operations prevent persistent identity from degenerating into an immutable mask.}
\label{fig:agentic-loop}
\end{figure}

Figure~\ref{fig:agentic-loop} summarizes this transition as a closed perception--memory--action loop. In the diagram, ``Update memory'' jointly represents fusion of new evidence and persistence of the revised carrier; the decomposition remains the five LAF decisions even when an implementation performs these operations in one module.

\subsection{Persistent identity is memory, not only a mask}

A reusable object identity lets an agent connect observations, language, and actions across time. OnlineAnySeg incrementally reconciles class-agnostic masks in a hash-map state, while OGScene3D and OnlinePG update Gaussian object or panoptic maps as new frames arrive \citep{tang_2025_onlineanyseg,zhu_2026_ogscene3d,zhai_2026_onlinepg}. These systems show why online segmentation cannot be evaluated only as frame throughput. A merge changes future association, retrieval, and correction; latency and state quality are coupled.

SAGOnline uses a different meaning of online: it constructs an explicit Gaussian label carrier from SAM2-propagated masks over rendered views, then provides immediate interaction and background mask refinement without fitting a scene-specific semantic field \citep{sun_2025_sagonline}. This is online query service over an initialized scene, not continual map maintenance under newly arriving physical observations. The distinction should remain visible whenever online segmentation, online mapping, and real-time rendering are compared.

Persistent state should therefore include more than the current label. An agent needs merge provenance, confidence, timestamps, supporting views, and possibly alternative identities. OGScene3D's confidence-aware revision is a step toward such memory \citep{zhu_2026_ogscene3d}. Append-only identity maps are efficient but poorly suited to correction after moved objects, temporary occlusion, or systematic foundation-model error. Selective rollback and local reassociation are likely more important than globally recomputing every feature.

DovSG makes selective repair concrete: it relocalizes the changed neighborhood, removes obsolete voxels, fuses new observations and regenerates affected graph edges rather than rebuilding the scene \citep{yan_2025_dovsg}. Its failures also clarify the requirement: local updates need explicit feature invalidation, because residual CLIP evidence can still direct an agent toward an object's previous location.

\subsection{Active observation closes the perception loop}

Passive methods accept the views supplied by a dataset. OpenNeRF instead locates regions where projected language features disagree, selects a novel camera, renders an image, and invokes the 2D encoder again \citep{engelmann_2024_opennerf}. Although the camera is virtual and the scene is static, this is an early form of semantic next-best-view selection. For a physical agent, the observation policy could include geometric uncertainty, expected identity disambiguation, semantic novelty, motion cost, and safety.

CADRNet makes this observation policy differentiable and query-conditioned: language predicts virtual camera configurations, CLIP encodes the resulting projections, and a 2D--3D grouping module feeds an MLLM mask decoder \citep{yu_2026_cadrnet}. Unlike OpenNeRF's uncertainty-triggered evidence acquisition, its views are optimized end-to-end for reasoning segmentation. Both approaches expose the same systems question: whether a focused view resolves uncertainty or merely reinforces an error already present in geometry, language, or the learned policy.

The important shift is that view selection becomes part of the segmentation algorithm and its cost. A system should be credited for the uncertainty it resolves, but charged for added sensing, rendering, and foundation-model inference. It should also be tested for confirmation bias: a pose selected from an incorrect reconstruction may reproduce the same occlusion or artifact. Active segmentation therefore requires calibrated uncertainty over geometry and semantics, not only a high relevance score.

REALM operationalizes this loop with an MLLM agent over a 3DGS feature field: global views establish context, local views disambiguate the target, and target-specific refinement turns the selected 2D identity into a 3D mask \citep{shi_2026_realm}. The ablation also exposes an agentic failure absent from a static lookup: increasing refinement from 50 to 500 or 1000 steps degrades the mask by overfitting selected views. An agent therefore needs a stopping rule grounded in held-out evidence, not simply permission to deliberate or optimize longer.

Fun3DU supplies a complementary task-first loop without mutating the carrier. Its LLM infers which contextual object contains or relates to the requested functional element, and that inference determines detection, view selection and fine-grained pointing before 3D lifting \citep{corsetti_2025_fun3du}. Compared with REALM's global-to-local refinement, this makes reasoning cheaper to isolate but also easier to audit: language parsing, context localization, point prompting and multi-view agreement form distinct checkpoints at which an agent can retain alternatives or ask for clarification.

\subsection{Relations expand the query interface}

Unary language features answer questions such as ``which points resemble a chair?'' Agents also need ``the chair beside the desk,'' containment, support, reachability, and free-space relations. PairGS persists pairwise grouping evidence and a cluster hierarchy; ReLaGS and OGScene3D attach named relations to Gaussian entities; Octree-Graph combines object semantics with adaptive occupancy and graph edges for retrieval and planning \citep{cha_2026_pairgs,xie_2026_relags,zhu_2026_ogscene3d,wang_2025_octree_graph}. These carriers make segmentation a substrate for structured reasoning rather than an endpoint.

Relation-capable systems need decomposed evaluation. Correct object masks do not guarantee correct edges, and correct language retrieval does not guarantee collision-free occupancy. One merge can alter several relations simultaneously. Future protocols should report unary segmentation, entity identity, relation accuracy, update consistency, and downstream task success separately, while tracing which carrier error caused the failure.

Open-World 3DSG extends this interface from graph lookup to retrieval-augmented multimodal reasoning \citep{yu_2026_ow3dsg}. Because object masks, best-view labels, VLM predicates, vector retrieval and final answers form a cascade, a plausible answer cannot substitute for node- and edge-level grounding evidence.

RelationField shows that an agent-facing carrier can answer pairwise affordance, support, composition and spatial queries directly from a field \citep{koch_2025_relationfield}. Such a carrier can identify which object matters to an action, but a persistent hallucinated predicate may be more damaging than a wrong unary label because it changes the inferred interaction between two entities.

\subsection{Query correction should be a first-class operation}

Current interfaces mostly accept a new category or prompt but not a correction to persistent state. Yet an agent may learn that two masks are one object, that a label was wrong, or that a previously static object has moved. External indices such as ExtrinSplat separate semantic hypotheses from geometry, and graph carriers localize some updates to nodes or edges \citep{ding_2026_extrinsplat,wang_2025_octree_graph}. These designs suggest a corrective interface: revise a name, split or merge an entity with provenance, invalidate supporting observations, and propagate only the consequences justified by the carrier.

Agentic 3D perception is therefore not synonymous with adding an LLM to a scene map. It requires state whose identities and uncertainty survive over time, an observation policy that can seek missing evidence, and update operations that distinguish geometry, topology, semantics, and relations. The five-stage decomposition remains useful because it reveals which stage a correction must revisit. EmbodiedSAM, ESAM++ and MoonSeg3R make this transition concrete by treating incoming masks as updates to persistent query state rather than as an offline batch \citep{xu_2025_embodiedsam,liu_2026_esampp,du_2026_moonseg3r}. Their progress is operational--real-time or monocular online inference--but a complete agentic memory still requires reversible identity decisions, confidence-aware forgetting and actions that acquire disambiguating observations.

\section{Framework-Derived Research Directions}

The literature has moved rapidly from feature projection to structured, online, and feed-forward carriers. The remaining problems are less about finding another place to store a CLIP vector and more about coordinating uncertainty, identity, generalization, and evaluation across the full pipeline.

\subsection{Joint uncertainty over geometry, association, and semantics}

Most systems estimate one uncertainty while treating the others as fixed. OpenNeRF measures cross-view feature variance; VALA estimates visibility-aware contribution; LEGaussians uses uncertainty to modulate spatial smoothing \citep{engelmann_2024_opennerf,wang_2025_vala,shi_2024_legaussians}. None alone represents the joint event that geometry is wrong, the projected support is ambiguous, the 2D teacher is unreliable, and several identities remain plausible. A useful carrier would preserve a factored uncertainty model and propagate it to segmentation, view selection, and correction. Calibration should be evaluated under pose noise, missing views, domain shift, and teacher disagreement, not only on in-distribution confidence.

\subsection{Dynamic and open-world identity maintenance}

Static reconstruction makes identity errors easier to hide because the carrier is built once. Online maps reveal that association is a sequence of state transitions \citep{tang_2025_onlineanyseg,zhai_2026_onlinepg,lee_2026_embodiedsplat}. Future systems should distinguish newly observed objects, reappearance after occlusion, actual motion, appearance change, and earlier association error. They need reversible merge histories, identity confidence, and bounded-memory strategies that do not erase long-tail entities. Benchmarks should include corrective queries and controlled changes rather than evaluating only the final aggregate map.

\subsection{Generalizable carriers without semantic collapse}

Per-scene fields fit detail but repeat optimization; feed-forward models amortize cost but may collapse unusual geometry or vocabulary. GOV-NeSF generalizes a semantic field, Uni3R and Ov3R jointly predict geometry and semantics with fewer geometric inputs, and Chorus distills multiple teachers into a reusable Gaussian encoder \citep{wang_2024_gov_nesf,sun_2026_uni3r,gong_2026_ov3r,li_2026_chorus}. HumanCrafter shows the strength and limitation of specialization: strong foundation features can support fast, consistent human-part reconstruction while the output vocabulary remains closed \citep{pan_2025_humancrafter}. Future work should report scene, geometry, vocabulary, and task generalization separately and study lightweight scene adaptation that retains rare structures instead of overwriting them.

PanSt3R and PartField extend amortization in complementary directions: the former predicts multi-view panoptic geometry without poses, whereas the latter learns a cross-shape hierarchy from inconsistent 2D and 3D part proposals \citep{zust_2025_panst3r,liu_2025_partfield}. Their remaining assumptions---keyframe/global mask selection for scenes, and object-scale canonical orientation for cross-shape use---show why feed-forward generalization still needs explicit stress tests rather than a single average benchmark.

SceneDINO and D-DITR expose two complementary research questions: how should an amortized model represent uncertainty over unseen 3D content, and how much image-grounded detail survives distillation into an image-free point model \citep{jevtic_2025_scenedino,abouzeid_2026_ditr}? Future evaluations should separate visible and completed regions and report the accuracy-cost frontier between multimodal teachers and deployable students.

\subsection{Carriers for overlap, hierarchy, and relations}

One vector or hard ID per primitive is poorly matched to overlapping affordances, materials, parts, and object groups. SAGA and GARField condition grouping on scale; PairGS stores pairwise evidence; ReLaGS and Octree-Graph add relational structure \citep{cen_2025_saga,kim_2024_garfield,cha_2026_pairgs,xie_2026_relags,wang_2025_octree_graph}. The open question is how to preserve this structure without making construction and querying quadratic. Sparse factor graphs, reversible cluster trees, external semantic indices, and query-dependent materialization are promising directions. Evaluation must include overlapping ground truth and relation-sensitive queries; a single flat mIoU cannot reward the intended capability.

\subsection{Native-3D benchmarks and complete cost reporting}

The field still mixes rendered masks, point labels, Gaussian assignments, proposal AP, and downstream navigation. A standard suite should provide the same scenes with native surface semantics, multi-view annotations, instance identities, part hierarchies, visibility, and repeated prompts. It should separate perception from oracle proposals and report first-query and repeated-query latency. Every runtime should specify reconstruction, 2D preprocessing, association, reconciliation, carrier construction, storage, and query. Energy or accelerator-hours are particularly important for amortized models whose deployment looks inexpensive only because training is omitted.

\subsection{A unified correspondence audit}

Many apparent semantic improvements may originate in different visibility thresholds, proposal coverage, or renderer weights. A unified correspondence audit would hold the 2D evidence and geometry fixed, then compare hard projection, depth-filtered projection, NeRF weights, Gaussian contribution, dominant-primitive registration, and learned association. It should measure not only final segmentation but support precision/recall, occlusion errors, boundary leakage, and sensitivity to pose/depth perturbation. LAF defines the variables and output contract required to make such an experiment interpretable; implementing the shared benchmark remains future work.

\subsection{Hybrid automatic and interactive carriers}

Find3D and PatchAlign3D demonstrate that scalable pseudo-part engines can yield fast reusable encoders, while GeoSAM2 shows that retaining a human-visible prompt preserves local control \citep{ma_2025_find3d,hadgi_2026_patchalign3d,deng_2026_geosam2}. A useful next step is a hybrid carrier that answers automatically from amortized memory, exposes pseudo-label provenance, and requests an image-space correction only where uncertainty warrants added interaction.

PartSLIP, COPS and CoSMo3D trace a useful amortization path from query-time boxes, through training-free dense feature clustering, to a reusable canonical point encoder \citep{liu_2023_partslip,garosi_2025_cops,jin_2026_cosmo3d}. The unresolved target is not merely faster inference: it is a carrier that retains COPS-like revisable decomposition and PartSLIP-like open prompts while achieving CoSMo3D-like direct deployment. Such a comparison needs the same rotated shapes, prompt sets and full data-engine ledger.

Across these directions, the recurring principle is to preserve revisable structure. Geometry, correspondence, identity, semantics, and relations should not become irreversibly entangled earlier than necessary. The best future carrier may not be the densest or most language-rich one, but the one that supports accurate queries, calibrated abstention, and local correction under a realistic cost budget.

Table 7 collects five empirical findings produced by the decision traces and states the conditions under which each remains valid. They are deliberately scoped to the audited operators, carrier contracts, evaluation domains, and cost components rather than asserted as universal laws.

\EvidenceTakeawaysTable

\section{Conclusion}

Transferring a 2D foundation model into 3D is not one lifting operation. It is a composition of decisions about generated evidence, geometric support, cross-view identity, semantic and granularity conflict, fusion, persistence, and query. LAF turns those decisions into an explicit trace and describes the resulting state through a carrier contract covering support, semantics, identity, uncertainty, provenance, and operations.

Applying the framework to 161 systems across six carrier families stress-tested its representation neutrality and exposed differences hidden by task- or representation-only taxonomies. Calibrated projection, renderer-mediated weights, learned correspondence, and temporary intermediaries can all implement association without deciding identity. Point attributes, continuous fields, Gaussian state, object inventories, relation graphs, and temporal memories can all persist segmentation while offering different query and correction contracts. Online, relational, dynamic, and feed-forward methods require feedback and state versioning, but they do not remove any of the five decisions.

The audit yields three operational rules. First, correspondence and identity must be reported separately: locating evidence on a surface does not determine which observations form an entity. Second, accuracy is meaningful only within a compatible output contract, and efficiency must include geometry, 2D preprocessing, association, reconciliation, carrier construction, storage, and first and repeated queries. Third, uncertainty and provenance must survive until the system no longer needs correction; an irreversible proposal, support error, or merge creates a ceiling that downstream language reasoning cannot remove.

The current validation is evidence-based rather than experimental: it establishes analytical coverage and saturation on the frozen corpus but does not yet provide prospective inter-annotator reliability or a controlled correspondence benchmark. Those limitations define the next tests of the framework. The broader design implication is nevertheless concrete. Persistent 3D perception for agents should be evaluated not only by the mask it returns, but by when it commits to identity, what evidence it retains, which revisions it supports, and what complete cost it incurs. The enduring problem is therefore not simply how to lift 2D knowledge into 3D, but how to make that knowledge coherent, accountable, and revisable.

\bibliographystyle{plainnat}
\bibliography{papers}
\end{document}

%% file: overview_figure.tex
\begin{figure*}[t]
\centering
\includegraphics[width=\textwidth]{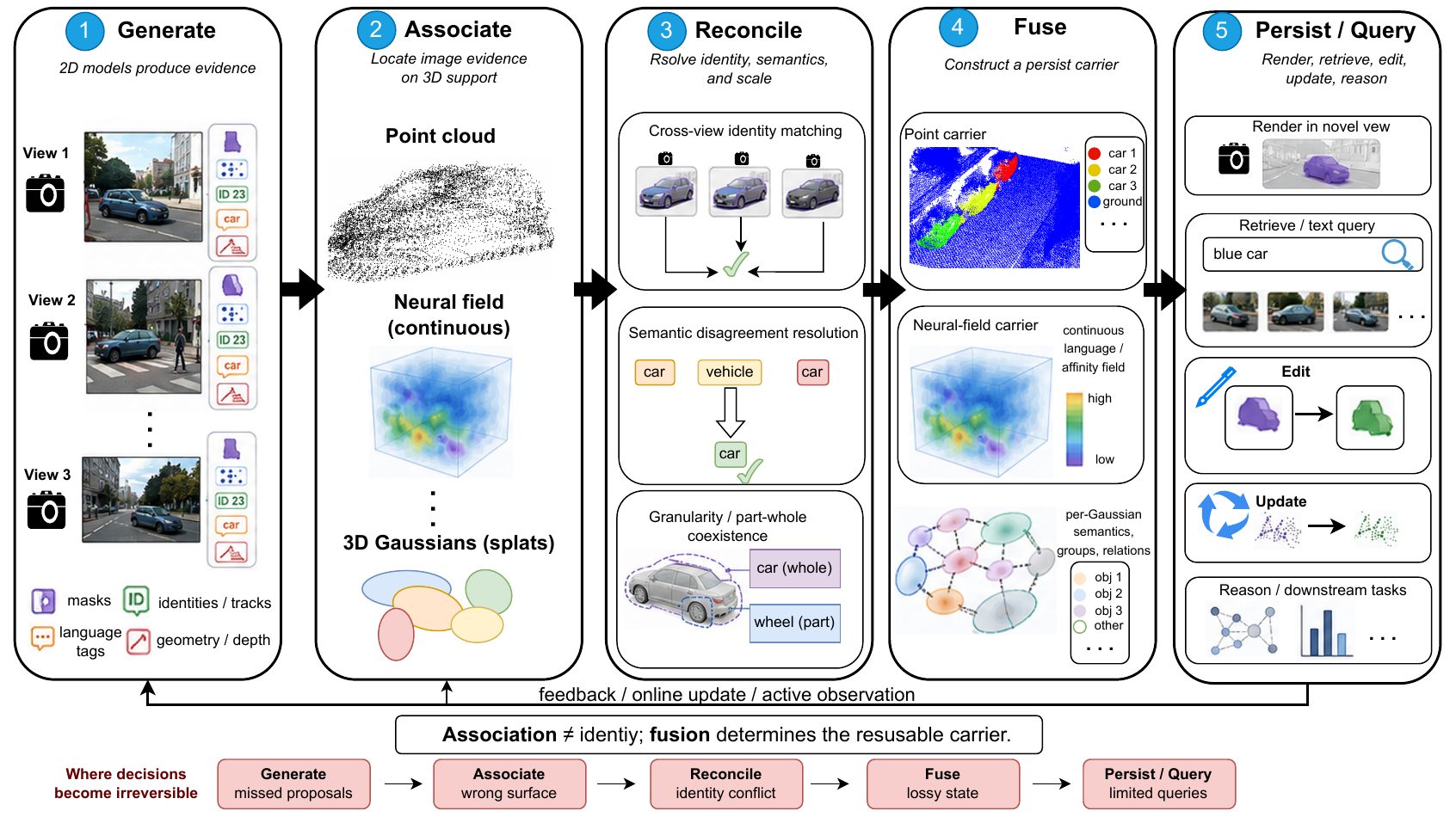}
\caption{Decision-centric LAF framework for 2D-to-3D foundation-model transfer. A system is represented by five operators---Generate, Associate, Reconcile, Fuse, and Persist/Query---that expose where image evidence is grounded in 3D, where cross-view identity, semantic disagreement, and granularity are resolved, and what persistent carrier is constructed. The feedback path represents online updates and active observation, while the lower row shows representative irreversible error propagation.}
\label{fig:pipeline}
\end{figure*}